\documentclass[letterpaper]{article} % DO NOT CHANGE THIS
\usepackage{aaai25}  % DO NOT CHANGE THIS
\usepackage{times}  % DO NOT CHANGE THIS
\usepackage{helvet}  % DO NOT CHANGE THIS
\usepackage{courier}  % DO NOT CHANGE THIS
\usepackage[hyphens]{url}  % DO NOT CHANGE THIS
\usepackage{graphicx} % DO NOT CHANGE THIS
\usepackage{natbib}  % DO NOT CHANGE THIS AND DO NOT ADD ANY OPTIONS TO IT
\usepackage{caption} % DO NOT CHANGE THIS AND DO NOT ADD ANY OPTIONS TO IT
\usepackage{subcaption}
\usepackage{float}
\usepackage{algorithm}

\usepackage{bm}

\usepackage{amsmath, amsthm, amssymb, amsfonts}
\usepackage{algpseudocode}

\usepackage{xcolor}

\theoremstyle{remark}

\newcommand{\ind}{\mathbb{I}}  
\newcommand{\Prob}{\mathbb{P}}
\newcommand{\E}{\mathbb{E}}
\usepackage{mathtools}

\usepackage{tikz}
\usepackage{pgfplots}
\pgfplotsset{compat=1.18} 
\usepackage{newfloat}
\usepackage{listings}
\usepackage{float}
\usepackage{booktabs}
\DeclareCaptionStyle{ruled}{labelfont=normalfont,labelsep=colon,strut=off} % DO NOT CHANGE THIS
\floatstyle{ruled}
\newfloat{listing}{tb}{lst}{}
\floatname{listing}{Listing}
\title{FairPOT: Balancing AUC Performance and Fairness with Proportional Optimal Transport}

\author {
    Pengxi Liu,
    Yi Shen,
    Matthew M. Engelhard,
    Benjamin A.	Goldstein,
    Michael J. Pencina,
    Nicoleta J.	Economou-Zavlanos,
    Michael M. Zavlanos
}

\affiliations {
    Duke University\\
    pengxi.liu@duke.edu, yi.shen478@duke.edu, m.engelhard@duke.edu, ben.goldstein@duke.edu,
    michal.pencina@duke.edu, nicoleta.economou@duke.edu,
    mz61@duke.edu
}

\usepackage{bibentry}
\nocopyright

\begin{document}
\maketitle
\begin{abstract}
% AUC-based fairness notions have gained increasing attention in high-stakes domains such as healthcare, finance, and criminal justice, as where fairness is evaluated over risk scores rather than binary outcomes. However, enforcing strict fairness often leads to significant degradation in AUC. In this work, we propose \textit{Proportional Optimal Transport} (PPOT), a model-agnostic post-processing framework that aligns score distributions across groups using optimal transport, but only partially—by transforming a controllable proportion (e.g., top-$\lambda$ quantile) of scores of disadvantaged group. Varying $\lambda$ enables an adjustable trade-off between xAUC disparities and AUC. We further extend PPOT to the partial AUC setting by integrating it with partial AUC optimization, allowing fairness interventions to focus on top-risk regions. Experiments on synthetic, public, and clinical datasets show that PPOT consistently outperforms existing post-processing methods in both global and partial AUC settings, often achieving improved fairness with little degration or even positive gains in utility. Its computational efficiency and practical flexibility make it well-suited for real-world deployment.

Fairness metrics utilizing the area under the receiver operator characteristic curve (AUC) have gained increasing attention in high-stakes domains such as healthcare, finance, and criminal justice. In these domains, fairness is often evaluated over risk scores rather than binary outcomes, and a common challenge is that enforcing strict fairness can significantly degrade AUC performance. To address this challenge, we propose Fair Proportional Optimal Transport (FairPOT), a novel, model-agnostic post-processing framework that strategically aligns risk score distributions across different groups using optimal transport, but does so selectively by transforming a controllable proportion, i.e., the top-$\lambda$ quantile, of scores within the disadvantaged group. By varying $\lambda$, our method allows for a tunable trade-off between reducing AUC disparities and maintaining overall AUC performance. Furthermore, we extend FairPOT to the partial AUC setting, enabling fairness interventions to concentrate on the highest-risk regions. Extensive experiments on synthetic, public, and clinical datasets show that FairPOT consistently outperforms existing post-processing techniques in both global and partial AUC scenarios, often achieving improved fairness with slight AUC degradation or even positive gains in utility. The computational efficiency and practical adaptability of FairPOT make it a promising solution for real-world deployment.

\end{abstract}


\section{Introduction}
Recent advancements in machine learning have raised equal excitement and concern, especially in high-stakes domains that involve a large extent of decision-making, such as healthcare \cite{rajkomar2018ensuring}, finance \cite{hardt2016equality}, and criminal justice \cite{ensign2018runaway}. A major concern lies in the biases embedded throughout the machine learning lifecycle \cite{suresh2021framework}. These biases may arise from various stages, including data collection, algorithm design, and model deployment, and can result in systematic discrimination against certain individuals or demographic groups \cite{mehrabi2021survey}. In this sense, fairness has emerged as a prominent topic in the study of machine learning ethics. Within this realm, researchers aim to address two fundamental questions: \textit{conceptually}, how to define and measure fairness, and \textit{methodologically}, how to achieve fairness \cite{pessach2022review}. 

A widely studied class of fairness notions is group fairness \cite{mehrabi2021survey}. It compares the outcomes of machine learning models across different groups defined by sensitive attributes, such as gender or race \cite{caton2024fairness}. Classical group fairness notions, such as demographic parity \cite{dwork2012fairness} and equalized odds \cite{hardt2016equality}, are framed around binary outcomes. They are typically evaluated using confusion matrices or statistical parity after thresholding continuous risk scores into binary predictions \cite{kallus2019fairness, caton2024fairness}.

However, in many real-world applications, especially in high-stakes settings, risk scores, i.e., continuous-valued outputs, are used directly to guide decisions. Risk scores convey the degree of uncertainty in model predictions and serve as intermediate outputs, rather than binary decisions. Unlike deterministic binary outcomes, unadjusted risk scores allow for more flexible and context-aware interventions \cite{kallus2019fairness}. For instance, in healthcare, a patient with a predicted risk score of 0.7 may receive priority compared to a patient with a risk score of 0.6, even if both are above a decision threshold. In this context, fairness concerns arise at the level of the scores themselves, motivating the need for methods that can address score-level group disparities, rather than just binary outcomes.

To this end, we focus on score-based group fairness. A representative metric in this setting is xAUC disparity \cite{kallus2019fairness}, which measures fairness from a bipartite ranking perspective. Specifically, xAUC measures the probability that risk scores rank randomly chosen positive samples from one group higher than the negative samples from another group, and vice versa. The xAUC disparity is then defined as the difference between these two probabilities, capturing the asymmetry in class distinguishability across groups.

To mitigate score-level disparities, various fairness-enhancing strategies have been proposed. Broadly, these can be categorized into pre-processing (modifying the input data), in-processing (changing the learning algorithm), and post-processing (adjusting model outputs) approaches in terms of the underlying stage at which fairness interventions are implemented \cite{pessach2022review, caton2024fairness}. In this work, we focus on post-processing methods, which are particularly appealing for their model-agnostic nature: they can be applied to any predictive model without modifying the training process or requiring access to internal parameters. This makes post-processing especially practical in real-world settings where models are already deployed or not easily modifiable. 

\begin{figure}
    \centering
    \includegraphics[width=0.8\linewidth]{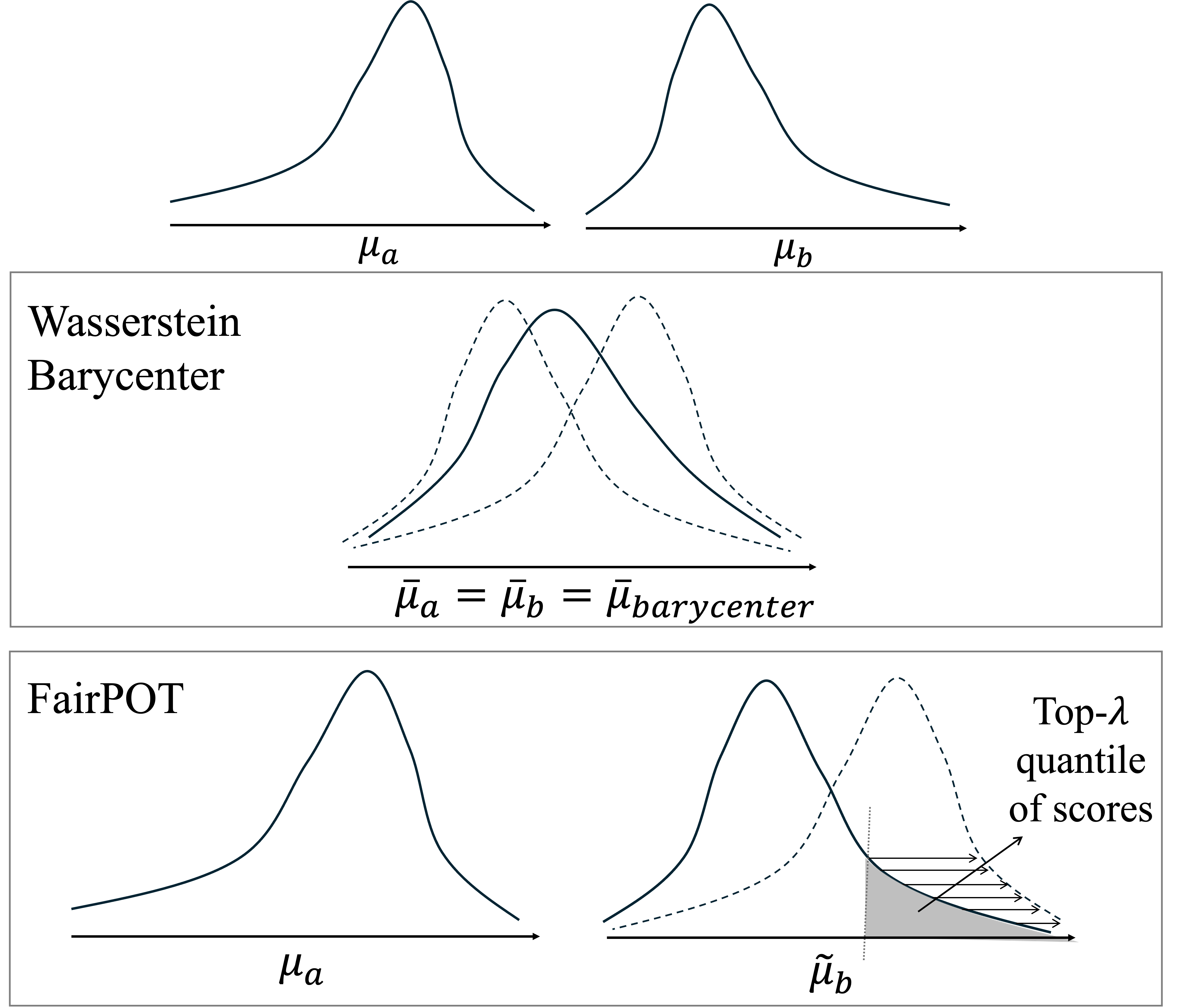}
    \caption{Comparison of FairPOT and Wasserstein barycenter-based methods.}
    \label{fig:wf_ppot}
\end{figure}

Among post-processing techniques, optimal transport has gained increasing attention as it provides a principled foundation for distribution alignment. Existing studies typically move score distributions from different groups toward a shared target, named as the Wasserstein barycenter \cite{jiang2020wasserstein, silvia2020general, xian2023fair}. As such, the post-processed scores follow identical distributions, reducing group-level discrepancies and promoting fairness at the score level. However, such repairs can introduce substantial deviations from the original scores, potentially harming model utility. 

In this work, we propose \textit{Fair Proportional Optimal Transport} (FairPOT), which, unlike related literature, applies optimal transport only to a proportion (e.g., top-$\lambda$ quantile) of the disadvantaged group, while keeping the advantaged group unchanged, as illustrated in Figure~\ref{fig:wf_ppot}. By limiting the adjustment to a selected portion of scores from the disadvantaged group, FairPOT ensures no modification to the advantaged group and provides precise control over the extent of alignment applied to the disadvantaged group. The parameter $\lambda$ allows for balancing fairness and accuracy—in our case, xAUC disparity and AUC. We further extend FairPOT to the partial AUC (pAUC) setting, where fairness and accuracy are evaluated within top-ranked regions of the score distribution, which are most critical in high-stakes decision-making scenarios. For instance, in medical screening tasks where the overall positive rate is low, it is often more important to ensure that high-risk individuals (i.e., those assigned high risk scores) are correctly prioritized, rather than focusing on the entire population. Our proposed method enables explicit balancing between fairness and accuracy in both global (AUC) and partial (pAUC) settings. We characterize the Pareto frontier of the fairness–accuracy plot as the set of solutions where no further improvement in fairness can be achieved without sacrificing accuracy, and vice versa. We validate our approach through extensive experiments on a variety of datasets, including synthetic, public, and real-world clinical data.

\section{Related Work}

\subsection{AUC-based Fairness}
AUC-based fairness metrics are particularly important for evaluating fairness of risk scores, as they assess ranking quality independently of thresholding. One early metric is pinned AUC \cite{dixon2018measuring}, which evaluates AUC on balanced datasets with equal subgroup representation. \citet{borkan2019nuanced} introduce three variants to capture groupwise ranking errors. \citet{kallus2019fairness} propose a notion named xAUC, measuring the probability that positive samples from one group are ranked above negative samples from another, with the xAUC disparity quantifying cross-group differences.

To reduce xAUC disparities, many methods approximate the non-differentiable AUC objective using surrogate losses \cite{yao2023stochastic}.  \citet{yang2023minimax} frame the AUC objective as a zero-sum game problem so as to utilize a stochastic gradient optimization algorithm. \citet{yao2023stochastic} approximate AUC objective and constraints with quadratic surrogate loss. However, such surrogate loss proxies may yield approximations that diverge significantly from the rank statistics they are intended to approximate \cite{rudin2018direct}.  To avoid this issue, \citet{cui2023bipartite} directly optimize bipartite rankings by greedily reordering cross-group instances to balance AUC and fairness, though at the cost of high computational complexity. Our FairPOT relates xAUC disparities with the divergence between score distributions of different groups, and applies optimal transport to reduce these disparities without retraining or complex reordering. 

\subsection{Partial AUC Optimization}
While AUC has been widely used to measure the discriminative ability of a scoring function, in certain scenarios, partial AUC, which measures partial area under ROC curve, can be a more practically informative metric \cite{dodd2003partial, carrington2020new}. Compared to AUC, partial AUC can be more appropriate for certain real-life scenarios \cite{narasimhan2013structural}, such as biomedical screening \cite{zhu2024improving} in clinical practice and Top-K ranking \cite{shi2024lower} in recommendation systems.

Optimization of partial AUC has also been thoroughly explored in the literature. Some early works use indirect methods to relate objectives to partial area under the ROC curve, rather than directly optimizing partial AUC, including p-norm push \cite{rudin2009p} and infinite push \cite{agarwal2011infinite, li2014top}. 
\citet{narasimhan2013structural, narasimhan2013svmpauctight} propose to use structural SVM to optimize the surrogate objective of partial AUC for linear models. \citet{rudin2018direct} employ mixed-integer programming to directly optimize partial AUC. Recent methods \cite{yang2021all, yao2022large, zhu2022auc} are based on stochastic gradient descent with approximate pairwise surrogate objectives which are amenable to deep learning. Despite growing interest in partial AUC, existing work has focused almost exclusively on model performance. Our work fills an important gap by incorporating fairness into partial AUC optimization, aiming to balance cross-group fairness and predictive performance in top-risk regions.

\subsection{Optimal Transport for Fairness}
Optimal transport has been widely used in the literature to achieve fairness \cite{gordaliza2019obtaining, jiang2020wasserstein, silvia2020general, zehlike2020matching, laclau2021all, buyl2022optimal, xian2023fair, han2024intra, ghassemi2025auditing}. Relevant methods span all three categories, i.e., pre-processing, in-processing, and post-processing. In pre-processing methods, optimal transport mainly aims to move features of different groups to their Wasserstein barycenter so as to minimize the deviation from original features \cite{gordaliza2019obtaining}. In-processing methods incorporate fairness into the training process by adding a regularization term that uses the Wasserstein distance to penalize differences between the score distributions of different groups \cite{buyl2022optimal}. Post-processing methods typically focus on score distributions of different groups. For instance, \citet{jiang2020wasserstein} propose to conduct quantile matching based on the Wasserstein-1 distance across different groups. Our FairPOT method partially aligns the scores across groups, allowing for flexible, model-agnostic adjustments to balance fairness and accuracy.

\subsection{Fairness-Accuracy Trade-off}
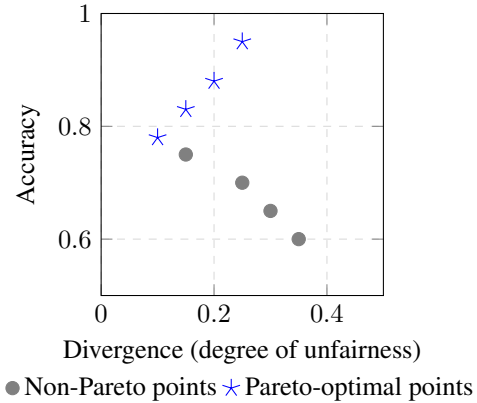
\begin{figure}[t]
\centering
\begin{tikzpicture}
\begin{axis}[
    width=0.3\textwidth,
    height=0.3\textwidth,
    xlabel={Divergence (degree of unfairness)},
    ylabel={Accuracy},
    xmin=0, xmax=0.5,
    ymin=0.5, ymax=1,
    grid=both,
    grid style={dashed, gray!30},
    legend style={at={(0.5,-0.25)}, anchor=north, legend columns=-1, draw=none},
    legend cell align={left},
]

% Non-Pareto points
\addplot[
    only marks,
    mark=*,
    color=gray,
    mark size=2.5pt
]
coordinates {
    (0.35,0.60)
    (0.30,0.65)
    (0.25,0.70)
    (0.15,0.75)
};

% Pareto points
\addplot[
    only marks,
    mark=star,
    mark options={solid},
    color=blue,
    mark size=3.5pt
]
coordinates {
    (0.10,0.78)
    (0.15,0.83)
    (0.20,0.88)
    (0.25,0.95)
};

\legend{Non-Pareto points, Pareto-optimal points}

\end{axis}
\end{tikzpicture}
\caption{Illustration of the Pareto frontier: Pareto-optimal points (blue stars) dominate non-Pareto points (gray dots).}
\label{fig:pareto_frontier}
\end{figure}

Alongside efforts to achieve fairness, recent studies have also explored the trade-off between fairness and accuracy \cite{kleinberg2016inherent}, motivating the need for mechanisms that balance both. A common approach leverages Pareto optimality \cite{mas1995microeconomic} to characterize solutions where improving one objective necessarily compromises another \cite{xiao2017fairness, martinez2020minimax, ge2022toward, wei2022fairness, xu2023fair}. Different studies interpret Pareto optimality from varying perspectives. \citet{martinez2020minimax} interpret this as ``no unnecessary harm” to any group. A more general formulation defines Pareto optimality in terms of the trade-off between fairness and accuracy \cite{xiao2017fairness, ge2022toward, wei2022fairness, xu2023fair}, requiring that no solution simultaneously worsens both metrics, as illustrated in Figure \ref{fig:pareto_frontier}. While our method is not derived from these prior approaches, we adopt the Pareto optimality principle to characterize the trade-off frontier between fairness and accuracy. Specifically, we construct the Pareto frontier by filtering out all dominated points based on AUC and xAUC disparity, and evaluate our method by comparing it against the non-dominated points achieved by other fairness intervention methods.

\section{Preliminaries}
Consider a probability space $(\Omega, \mathcal{F}, \mathbb{P})$, and let $(X, Y, G): \Omega \rightarrow \mathcal{X} \times\{0,1\} \times\{a, b\}$ be a triplet of random variables defined on this space. We denote by $\mathcal{D}$ the joint distribution of $(X, Y, G)$ induced by $\mathbb{P}$, i.e., $(X, Y, G) \sim \mathcal{D}$. Here, $X \in \mathbb{R}^d$ is the feature vector, $Y \in\{0,1\}$ is the binary label, and $G \in\{a, b\}$ is the sensitive attribute representing demographic groups where we want to impose fairness interventions. We focus on the case where the sensitive attribute $G$ is binary, i.e., group $a$ or $b$, which is common in many applications such as gender (e.g., male vs. female) or race (e.g., White vs. non-White). While sensitive attributes may naturally take on multiple categories, the binary setting allows for simpler group-wise comparisons and is widely adopted in the fairness literature. Denote a scoring function $f:\mathcal{X}\to [0,1]$, where for each $x\in\mathcal{X}$, $f(x)$ is a real-valued score indicating the model belief of being labeled positively given $x$, i.e., $f(x): = \Prob[Y=1 \mid X=x]$ \cite{jiang2020wasserstein}. We also denote $\ind[\cdot]$ as the indicator function.

\subsection{AUC and AUC-based Fairness}

The AUC of a given score function $f$ represents the probability that $f$ assigns a strictly higher score to a randomly chosen positive instance than to a randomly chosen negative instance, which is
\begin{equation*}
    \mathrm{AUC}(f) \;=\; \Prob\bigl[f(X^+) > f(X^-)\bigr],
\end{equation*}
where $X^+$ and $X^-$ are independently drawn from $\mathcal{D}$ conditioning on $Y=1$ and $Y=0$, respectively. Equivalently, we can rewrite the $\mathrm{AUC}(f)$ as follows
\begin{equation*}
  \mathrm{AUC}(f)
    \;=\;
  \E\Bigl[\ind\bigl\{f(X^+) > f(X^-)\bigr\}\Bigr],
\end{equation*}
which calculates the fraction of positive-negative pairs that are correctly ranked by $f$. Consider now a finite sample dataset $\{(x_i, y_i, \hat{s}_i)\}_{i=1}^N$, where $\hat{s}_i=f(x_i)$, and denote the index sets
\[
  \mathcal{I}^+ = \{i \mid y_i=1\},
  \quad
  \mathcal{I}^- = \{j \mid y_j=0\}.
\]
Then the empirical AUC, $\widehat{\mathrm{AUC}}$, is defined by
\begin{equation*}
  \widehat{\mathrm{AUC}}(f)
    =
  \frac{
    \sum_{i\in \mathcal{I}^+}\!\sum_{j\in \mathcal{I}^-}
    \ind\bigl[\hat{s}_i > \hat{s}_j\bigr]
  }{
    |\mathcal{I}^+|\;\cdot\;|\mathcal{I}^-|
  }.
\end{equation*}
We assume that $|\mathcal{I}^+|>0$ and $|\mathcal{I}^-|>0$; otherwise, $\widehat{\mathrm{AUC}}(f)$ is set to zero to indicate that no valid positive-negative comparisons can be made.

While AUC is widely used to evaluate the discriminative ability of a scoring function, it does not account for sensitive attributes or disparities between groups. In line with the metric proposed by \citet{kallus2019fairness}, we use xAUC as the metric to measure how well $f$ ranks positives in one group above negatives in another group. We refer to this quantity as the (cross-group) xAUC, highlighting that it compares samples across different demographic groups. Define
\begin{equation*}
  \mathrm{xAUC}_{a\to b}(f)
    \;=\;
  \Prob\bigl[f(X_a^+)>f(X_b^-)\bigr],
\end{equation*}
where $X_a^+$ and $X_b^-$ are independently drawn from $\mathcal{D}$ conditioning on the events $\{Y=1,G=a\}$ and $\{Y=0,G=b\}$, respectively. Analogously, we define $\mathrm{xAUC}_{b\to a}(f)$ as the probability that a positive instance from group $b$ receives a higher score than a negative instance from group $a$, i.e.,
\begin{equation*}
  \mathrm{xAUC}_{b\to a}(f)
    \;=\;
  \Prob\bigl[f(X_b^+)>f(X_a^-)\bigr].
\end{equation*}
To quantify the degree of fairness across groups, we define the disparity between xAUC as
\begin{equation*}
  \Delta\mathrm{xAUC}(a,b)  =\bigl|\mathrm{xAUC}_{a\to b}(f)
        -
        \mathrm{xAUC}_{b\to a}(f)\bigr|.
\end{equation*}
Specifically, we say that xAUC-based fairness is achieved with a tolerance degree of $\varepsilon \ge 0$ if $\Delta\mathrm{xAUC}(a,b)\le \varepsilon$ for a given $\varepsilon \ge 0$ \cite{yao2023stochastic}. Empirically, let \[\mathcal{I}_a^+ = \{\,i\mid y_i=1, g_i=a\}, \quad \mathcal{I}_b^- = \{\,j\mid y_j=0, g_j=b\}\] denote the sets of positive samples from group $a$ and negative samples from group $b$, respectively. 
Then the empirical xAUC is defined as
\begin{equation*}
  \widehat{\mathrm{xAUC}}_{a\to b}(f)
    =
  \frac{
    \sum_{i\in \mathcal{I}_a^+}
    \sum_{j\in \mathcal{I}_b^-}
    \ind[\hat{s}_i > \hat{s}_j]
  }{
    |\mathcal{I}_a^+|\;\cdot\;|\mathcal{I}_b^-|
  }.
\end{equation*}
Hereby, we also assume that $\left|\mathcal{I}_a^{+}\right|>0$ and $\left|\mathcal{I}_b^{-}\right|>0$; otherwise, $\widehat{\mathrm{xAUC}}_{a \rightarrow b}(f)$ is set to zero to reflect the absence of valid pairwise comparisons.
Similarly, $\widehat{\mathrm{xAUC}}_{b \rightarrow a}(f)$ is defined by swapping $a$ and $b$. The empirical cross-group xAUC disparity is then given by
\begin{equation*}
  \Delta\widehat{\mathrm{xAUC}}(a,b)  =\bigl|\widehat{\mathrm{xAUC}}_{a\to b}(f)
        -
        \widehat{\mathrm{xAUC}}_{b\to a}(f)\bigr|.
\end{equation*}

\subsection{Partial AUC and Partial AUC-based Fairness}
While a high AUC reflects good global separation of positive and negative instances, it does not necessarily imply strong ability to distinguish between classes within localized regions of the score distribution. A model may attain a high AUC by correctly ranking many instances in the middle range, while still failing to separate positives and negatives among the top-scoring individuals—who are often critical in high-stakes applications. For instance, in medical diagnosis, individuals with higher predicted risk scores require more attention and prioritization.

Partial AUC is a refinement of AUC that focuses on specific regions of the ROC curve that are practically more important for decision-making. Popular definitions of partial AUC fall into two categories: (i) restricting attention to a specified range of false positive rates (FPR) \cite{dodd2003partial, narasimhan2013structural, iwata2020semi, zhu2022auc}, and (ii) simultaneously restricting both FPR and true positive rates (TPR) \cite{yang2019two, yang2021all, chaibub2024novel, shi2024lower}. For example, \citet{dodd2003partial} define partial AUC as the area under the ROC curve restricted to $\mathrm{FPR} \leq \alpha$, under the rationale that in some domains (e.g., medical diagnosis), we cannot afford a high false alarm rate.

However, it is often unclear how to choose these ranges of FPR or TPR. For example, in a medical screening task where the disease rate is 3\%, doctors may want to focus on the top 10\% of patients with the highest risk scores—about three times the base rate. This top percentage is easier to decide in practice, while FPR or TPR values are harder to connect to real needs. Motivated by this, we propose a new notion of partial AUC defined over the top-$\alpha$ scoring region. Rather than integrating over FPR/TPR, we restrict attention to instances whose predicted scores fall within the top-$\alpha$ quantile of the score distribution.

Denote the $\alpha$-quantile of the score distribution induced by $f$ as $t_\alpha$, i.e., $\Prob[f(X) \ge t_\alpha] = \alpha$ with $X$ drawn from the marginal distribution of $\mathcal{D}$ over features. We consider independent random draws \(X^{+} \sim \mathcal{D}_1\) and \(X^{-} \sim \mathcal{D}_0\), where $\mathcal{D}_1$ and $\mathcal{D}_0$ denote the conditional distributions of $X$ given $Y=1$ and $Y=0$, respectively. Based on the joint sampling of \((X^+, X^-)\), the partial AUC over the top-$\alpha$ region is defined as
\begin{align*}
  & p\mathrm{AUC}(f;\alpha) \\
  &= \Prob\Bigl[f(X^+) > f(X^-) \mid f(X^+)\ge t_\alpha, f(X^-)\ge t_\alpha\Bigr].
\end{align*}
In other words, it measures how well the score function distinguishes positives from negatives among the top-scoring instances. When $\Prob[f(X^-)\ge t_\alpha] = 0$, we set pAUC to 1; when $\Prob[f(X^+)\ge t_\alpha] = 0$, we set it to 0, reflecting perfect or failed separation in the top-$\alpha$ region.
%
% For example, in healthcare or screening tasks, limited resources often mean that only patients with the highest predicted risks receive further examinations or treatments. In such settings, it is important that the model remains good performance within this high-risk subset, rather than only across the entire population. 
%
Empirically, let the predicted scores be $\boldsymbol{\hat{s}} = (\hat{s}_1, \hat{s}_2, \ldots, \hat{s}_N)$ sorted in descending order, and define $N_\alpha = \lceil \alpha N \rceil$. Let \(\mathcal{I}_{N_\alpha}^+ = \{i \mid y_i=1,\; i \le N_\alpha\}\) and \(\mathcal{I}_{N_\alpha}^- = \{j \mid y_j=0,\; j \le N_\alpha\},\) with $N_\alpha^+ = |\mathcal{I}_{N_\alpha}^+|$ and $N_\alpha^- = |\mathcal{I}_{N_\alpha}^-|$. Then, the empirical partial AUC is defined as
\begin{align*}
  \widehat{p\mathrm{AUC}}(f; \alpha)
  = \frac{
    \sum_{i\in \mathcal{I}_{N_\alpha}^+} \sum_{j\in \mathcal{I}_{N_\alpha}^-} \ind[\hat{s}_i > \hat{s}_j]
  }{
    N_\alpha^+ \cdot N_\alpha^-
  }.
\end{align*}
Alternatively, it can be written as
\begin{align*}
  \widehat{p\mathrm{AUC}}(f; \alpha)
  = \frac{
    \sum_{i\in \mathcal{I}^+} \sum_{j\in \mathcal{I}^-} 
    \ind[\hat{s}_i > \hat{s}_j] \; \ind[\hat{s}_j \ge \hat{s}_{N_\alpha}]
  }{
    N_\alpha^+ \cdot N_\alpha^-
  },
\end{align*}
where $\hat{s}_{N_\alpha}$ denotes the $N_\alpha$-th largest score. We also assume that $N_\alpha^+ > 0$ and $N_\alpha^- > 0$. Similarly, pxAUC-based fairness is considered satisfied with tolerance $\varepsilon \ge 0$ if $\Delta p\mathrm{xAUC}(a,b) \le \varepsilon$,

We now focus on fairness among the top-ranked instances. Let $X_a^+$ and $X_b^-$ be drawn from $\{X \mid Y=1, G=a\}$ and $\{X \mid Y=0, G=b\}$, respectively, both restricted to $f(\cdot) \ge t_\alpha$. Define
\begin{align*}
  & p\mathrm{xAUC}_{a\to b}(f;\alpha) \\
  &= \Prob\bigl[f(X_a^+) > f(X_b^-) \mid f(X_a^+) \ge t_\alpha,\, f(X_b^-) \ge t_\alpha\bigr], \\
  & p\mathrm{xAUC}_{b\to a}(f;\alpha) \\
  &= \Prob\bigl[f(X_b^+) > f(X_a^-) \mid f(X_b^+) \ge t_\alpha,\, f(X_a^-) \ge t_\alpha\bigr],
\end{align*}
and the partial fairness gap as
\begin{align*}
  \Delta p\mathrm{xAUC}(a,b;\alpha) = |p\mathrm{xAUC}_{a\to b}(f;\alpha) - p\mathrm{xAUC}_{b\to a}(f;\alpha)|.
\end{align*}
Empirically, for the top-$N_\alpha$ instances in each group, define index sets $\mathcal{I}_{N_\alpha,a}^+ = \{i \mid y_i=1, g_i=a, i \le N_\alpha\}$ and $\mathcal{I}_{N_\alpha,b}^- = \{j \mid y_j=0, g_j=b, j \le N_\alpha\}$. Then, the empirical partial cross-group xAUCs are computed as
\begin{align*}
  \widehat{p\mathrm{xAUC}}_{a\to b}(f;\alpha)
  = \frac{\sum_{i \in \mathcal{I}_{N_\alpha,a}^+} \sum_{j \in \mathcal{I}_{N_\alpha,b}^-} \ind[\hat{s}_i > \hat{s}_j]}{|\mathcal{I}_{N_\alpha,a}^+| \cdot |\mathcal{I}_{N_\alpha,b}^-|},
\end{align*}
with $\widehat{p\mathrm{xAUC}}_{b\to a}(f;\alpha)$ defined analogously. We also assume $|\mathcal{I}_{N_\alpha,a}^+|>0$ and $|\mathcal{I}_{N_\alpha,b}^-|>0$; if not, the corresponding empirical estimates are set to zero. The corresponding empirical partial xAUC disparity is
\begin{align*}
  \Delta \widehat{p\mathrm{xAUC}}(a,b;\alpha) = |\widehat{p\mathrm{xAUC}}_{a\to b}(f;\alpha) - \widehat{p\mathrm{xAUC}}_{b\to a}(f;\alpha)|.
\end{align*}

\subsection{Optimal Transport}
In this work, we focus on optimal transport in discrete settings \cite{monge1781memoire, kantorovich1942transfer}. Specifically, we consider two empirical measures
\[
\mu = \sum_{i=1}^n u_i \delta_{z_i}, \quad \nu = \sum_{j=1}^m v_j \delta_{z_j'},
\]
where $\delta_{z}$ denotes the Dirac delta measure at location $z$, $\{z_i\}_{i=1}^n$ and $\{z_j'\}_{j=1}^m$ are the support points, and $u, v$ are their associated probability mass vectors (typically uniform). The objective is to find a coupling $\gamma \in \mathbb{R}^{n \times m}$ that transports $\mu$ to $\nu$ while minimizing the total transport cost. Formally, the discrete optimal transport problem solves
\begin{equation}\label{eqn:optimal_transport_gamma}
\gamma^* = \arg\min_{\gamma \in \Gamma(\mu, \nu)} \sum_{i=1}^n \sum_{j=1}^m \gamma_{i,j} C_{i,j},
\end{equation}
where $\Gamma(\mu,\nu) = \{ \gamma \ge 0 \mid \gamma \mathbf{1}_m = u, \; \gamma^\top \mathbf{1}_n = v \}$ is the set of couplings with prescribed marginals, and $C_{i,j} = c(z_i, z_j')$ defines the cost of transporting mass from $z_i$ to $z_j'$. In our setting, we use the squared Euclidean distance as the cost, i.e., $C_{i,j} = (z_i - z_j')^2$. Once the optimal coupling $\gamma^*$ is found, we compute the barycentric projection (e.g., \citet{seguy2017large}) by
\begin{equation}\label{eqn:projection}
\tilde{z}_i = n\sum_{j=1}^m \gamma_{i,j}^* z_j', \quad \forall i \in \{1, \dots, n\}.
\end{equation}
The barycentric projection defines a mapping from $z_i$ toward $z_j'$ based on the optimal coupling $\gamma^*$.

\section{Fair Proportional Optimal Transport (FairPOT)}
\label{sec:auc-fairness}
Existing studies focusing on AUC fairness \cite{fong2021fairness, yang2023minimax, yao2023stochastic, cui2023bipartite} face a fundamental challenge: enforcing AUC fairness often leads to a deterioration in overall AUC performance. To address this challenge, we propose Fair Proportional Optimal Transport (FairPOT), a model agnostic post-processing method that employs optimal transport to align score distributions across different demographic groups to achieve fairness in terms of AUC and partial AUC with a tuning parameter $\lambda\in[0,1]$ to control the fairness-vs-accuracy trade-off.

\subsection{Problem Definition}
\begin{figure*}
    \centering
    \includegraphics[width=0.9\linewidth]{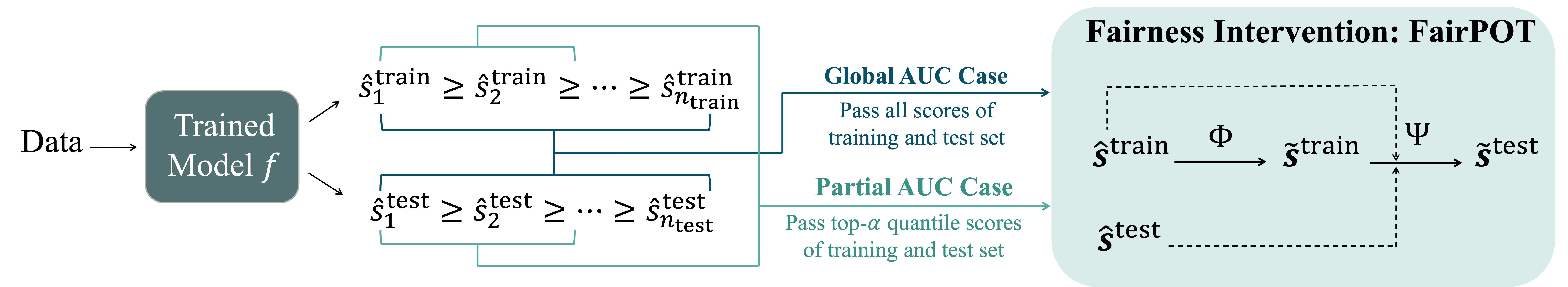}
    \caption{FairPOT framework. The partial transport \(\Phi\) (Eq.~\eqref{phi}) maps training scores of group \(b\) to \(\tilde{\boldsymbol{s}}_b^{\lambda,\text{train}}\), which are then used in the interpolation mapping \(\Psi\) (Eq.~\eqref{psi}) to adjust test scores.
}
    \label{fig:framework_partial}
\end{figure*}

We study two settings where FairPOT can be applied: the standard (global) AUC setting and the partial AUC setting. The key difference lies in the scope of scores used for fairness intervention and evaluation as illustrated in Figure \ref{fig:framework_partial}. In the global case, the entire score distribution is considered, whereas in the partial case, only the top-ranked (e.g., top-$\alpha$ quantile) region is evaluated and adjusted. In both cases, our goal is to trace the trade-off curve between fairness and accuracy by partially aligning scores of the disadvantaged group via optimal transport. Note that our fairness intervention is model-agnostic—it can be applied to any trained model, including black-box predictors. Our definition of the disadvantaged group and its relation to score distributions is further discussed in the Appendix. 

\paragraph{Global AUC Case.} Recall that we focus on binary sensitive attributes $G \in {a,b}$. Let $\boldsymbol{\hat{s}}_a = (\hat{s}_{a,1}, \ldots, \hat{s}_{a,n_a})$ and $\boldsymbol{\hat{s}}_b = (\hat{s}_{b,1}, \ldots, \hat{s}_{b,n_b})$ denote the predicted score vectors for groups $a$ and $b$, respectively, obtained from scoring function $f$. Each element $\hat{s}_{a,i} = f(x_i) \in [0,1]$ represents the predicted risk score for the $i$-th instance in group $a$, and similarly for group $b$.

We use $\widehat{\mathrm{AUC}}(f)$ and $\Delta\widehat{\mathrm{xAUC}}(a,b)$ to measure predictive performance and fairness. Our objective is to construct a parametric family of partially transported scores $ \{\boldsymbol{\tilde{s}}_b^\lambda\}_{\lambda \in [0,1]} $, where $\lambda$ controls the degree of intervention to balance these two goals. Specifically, as $\lambda$ varies, our goal is to obtain a trade-off curve in the $(\Delta\widehat{\mathrm{xAUC}}, \widehat{\mathrm{AUC}})$ space.

\paragraph{Partial AUC Case.} In many real-world applications, such as screening or triage, decisions are based on only a small fraction of individuals with the highest predicted scores. To reflect this, we consider evaluating both performance and fairness within the top-$\alpha$ region of the score distribution.

Let $\alpha \in [0,1]$ denote the fraction of top-ranked instances used for evaluation. We compute the partial AUC $\widehat{p\mathrm{AUC}}(f;\alpha)$ and partial xAUC disparity $\Delta \widehat{p\mathrm{xAUC}}(a,b;\alpha)$ over this region. Similar to the global case, our objective is to construct a series of transformed score vectors $\boldsymbol{\tilde{s}}_b^{\lambda,\alpha}$ by applying optimal transport only within the top-$\alpha$ subset of scores. As $\lambda$ varies, our goal is to produce a family of adjusted scores that define a trade-off curve in the $(\Delta \widehat{p\mathrm{xAUC}}, \widehat{p\mathrm{AUC}})$ space. This partial-view analysis reflects realistic constraints where interventions are focused on high-risk populations.

\subsection{Method}
\paragraph{Global AUC Case.} We apply optimal transport to align the distributions of scores of different groups. 
%In practice, we often transform only score distribution from one group (often the disadvantaged group) to another group to reduce cross-group disparities. 
Without loss of generality, we denote $a$ as the advantaged group and $b$ as the disadvantaged group. 
We treat the predicted score vectors $\boldsymbol{\hat{s}}_a$ and $\boldsymbol{\hat{s}}_b$ as support points of the empirical score distributions. 
Specifically, we define the empirical marginals as
$\mu_a = \frac{1}{n_a} \sum_{i=1}^{n_a} \delta_{\hat{s}_{a,i}}, 
\mu_b = \frac{1}{n_b} \sum_{j=1}^{n_b} \delta_{\hat{s}_{b,j}},$
where $n_a$, $n_b$ are the number of samples in groups $a$ and $b$, respectively. The optimal transport plan $\gamma^* \in \Gamma(\mu_b, \mu_a)$ is obtained by solving \eqref{eqn:optimal_transport_gamma}. 
% \[
% \gamma^* = \arg\min_{\gamma \in \Gamma(\mu_b, \mu_a)} \sum_{i,j} \gamma_{i,j} C_{i,j},
% \]
% where $C_{i,j} = \| \hat{s}_{b,i} - \hat{s}_{a,j} \|^2$ denotes the squared Euclidean distance between scores. 
Once the optimal plan $\gamma^*$ is obtained, we compute the transported scores $\boldsymbol{\tilde{s}}_b$ by applying the barycentric projection defined in \eqref{eqn:projection}.

Applying the full transport map moves the scores of group $b$ onto the support points of group $a$, such that each adjusted score of group $b$ lies within the convex hull of scores from group $a$. This alignment substantially reduces the distributional gap between group $a$ and group $b$, compared to the original unadjusted scores. Intuitively, by transporting each score in group $b$ toward the support of group $a$, the two distributions become more closely aligned in shape and range. This reduces mismatches in score comparisons between groups, which helps improve symmetry in pairwise rankings. Symmetry ensures that positives from one group are not consistently ranked below negatives from another. However, altering the score distribution in this manner may distort the global ordering between positive and negative instances, leading to potential degradation in overall AUC. For example, a positive instance from the advantaged group may initially have a higher score than a negative instance from the disadvantaged group, but after transport, the adjusted score of the negative instance may exceed that of the positive one—resulting in incorrect ranking and reduced AUC. To flexibly balance the trade-off between fairness and accuracy, we introduce a parameter $\lambda \in [0,1]$ that controls the fraction of scores to be transported. Specifically, we sort the original scores $\boldsymbol{\hat{s}}_b$ in descending order and formally define the partial transport mapping $\Phi$ as
\begin{equation}
\label{phi}
\Phi: (\boldsymbol{\hat{s}}_b; \gamma^*, \lambda) \longmapsto \boldsymbol{\tilde{s}}_b^\lambda,
\end{equation}

where
\[
\tilde{s}_{b,i}^\lambda =
\begin{cases}
n_b \sum\limits_{j=1}^{n_a} \gamma_{ji}^* \, \hat{s}_{a,j}, & \forall i \in \mathcal{I}_{b,\lambda}^{(\mathrm{top})},\\
\hat{s}_{b,i}, & \forall i \in \{1, \ldots, n_b\} \text{ and } i \notin \mathcal{I}_{b,\lambda}^{(\mathrm{top})},
\end{cases}
\]
 and $
n_{b,\lambda} = \lceil \lambda \, n_b \rceil, \quad \mathcal{I}_{b,\lambda}^{(\mathrm{top})} = \left\{ i : \hat{s}_{b,i} \leq \hat{s}_{b,n_{b,\lambda}} \right\}.
$ 

\begin{algorithm}[t]
\caption{FairPOT: Global AUC Case.}
\label{alg:full_ppot}
\begin{algorithmic}[1]
\State \textbf{Input:} Predicted scores $\boldsymbol{\hat{s}}^{\text{train}}$, $\boldsymbol{\hat{s}}^{\text{test}}$; group labels $\boldsymbol{g}^{\text{train}}$, $\boldsymbol{g}^{\text{test}}$; a set of discrete trade-off parameters $\Lambda = \{\lambda_1, \lambda_2, \dots, \lambda_L\} \subseteq [0,1]$.
\State \textbf{Output:} Transported scores $\boldsymbol{\tilde{s}}^{\lambda, \text{train}}$, $\boldsymbol{\tilde{s}}^{\lambda, \text{test}}$ for each $\lambda \in \Lambda$; Pareto frontier.
\State Sort $(\boldsymbol{\hat{s}}^{\text{train}}, \boldsymbol{g}^{\text{train}})$ in descending order of $\boldsymbol{\hat{s}}^{\text{train}}$.
\State Sort $(\boldsymbol{\hat{s}}^{\text{test}}, \boldsymbol{g}^{\text{test}})$ in descending order of $\boldsymbol{\hat{s}}^{\text{test}}$.
\State Extract group-specific scores
$
\boldsymbol{\hat{s}}_a^{\text{train}} \gets \{\hat{s}_i^{\text{train}} : g_i^{\text{train}} = a\}$ and
$
\boldsymbol{\hat{s}}_b^{\text{train}} \gets \{\hat{s}_i^{\text{train}} : g_i^{\text{train}} = b\}
$.
\State Define empirical distributions $\mu_a^{\text{train}} \gets \frac{1}{n_a} \sum_{i=1}^{n_a} \delta_{\hat{s}_{a,i}^{\text{train}}}$ and $\mu_b^{\text{train}} \gets \frac{1}{n_b} \sum_{j=1}^{n_b} \delta_{\hat{s}_{b,j}^{\text{train}}}$.
\State Define cost matrix $C_{i,j} \gets(\hat{s}_{b,i}^{\text{train}} - \hat{s}_{a,j}^{\text{train}})^2$.
\State Compute optimal transport plan
\[\gamma^* \gets \arg\min_{\gamma \in \Gamma(\mu_b^{\text{train}}, \mu_a^{\text{train}})} \sum_{i,j} \gamma_{i,j} C_{i,j}.\]
\For{each $\lambda \in \Lambda$}
    \State Transport train scores
    $\boldsymbol{\tilde{s}}_b^{\lambda, \text{train}} \gets \Phi(\boldsymbol{\hat{s}}_b^{\text{train}}; \gamma^*, \lambda)$.
    \State Map test scores
    $\boldsymbol{\tilde{s}}_b^{\lambda, \text{test}} \gets \Psi(\boldsymbol{\hat{s}}_b^{\text{train}}; \boldsymbol{\tilde{s}}_b^{\lambda, \text{train}}, \boldsymbol{\hat{s}}_b^{\text{test}})$.
    \State Construct full transported scores
    \[
    \boldsymbol{\tilde{s}}^{\lambda, \text{train}} \gets \text{Merge}(\boldsymbol{\hat{s}}_a^{\text{train}}, \boldsymbol{\tilde{s}}_b^{\lambda, \text{train}}),
    \]
    \[
    \boldsymbol{\tilde{s}}^{\lambda, \text{test}} \gets \text{Merge}(\boldsymbol{\hat{s}}_a^{\text{test}}, \boldsymbol{\tilde{s}}_b^{\lambda, \text{test}}).\]

    \State Evaluate
    $
    \widehat{\mathrm{AUC}}, \Delta\widehat{\mathrm{xAUC}}
    $ on $\boldsymbol{\tilde{s}}^{\lambda, \text{test}}$.
\EndFor
\State Identify non-dominated pairs of $(\Delta\widehat{\mathrm{xAUC}}, \widehat{\mathrm{AUC}})$ to form Pareto frontier across all $\lambda \in \Lambda$.

\end{algorithmic}
\end{algorithm}

As $\lambda$ varies from 0 to 1, we obtain a family of post-processed score vectors $\boldsymbol{\tilde{s}}_b^\lambda$, interpolating between no transport ($\lambda=0$) and full transport ($\lambda=1$). For each $\lambda \in [0,1]$, we evaluate the corresponding empirical AUC and fairness metrics based on the transformed scores $\boldsymbol{\tilde{s}}_b^\lambda$. Specifically, we compute $\bigl(\widehat{\mathrm{AUC}}(\boldsymbol{\tilde{s}}_b^\lambda),\; \Delta\widehat{\mathrm{xAUC}}(a,b; \boldsymbol{\tilde{s}}_b^\lambda)\bigr),$
which traces a curve in the fairness–accuracy plane as $\lambda$ varies. To formally characterize the optimal trade-offs, we adopt the principle of Pareto optimality. A point is Pareto-optimal if no other point achieves both higher AUC and lower fairness gap, with strict improvement in at least one. In practice, we identify the Pareto frontier by removing any point that is dominated by another, meaning there exists a point with better AUC and better fairness. The resulting curve captures the best achievable balance between AUC and fairness for different levels of transport.

Note that the post-processed scores are only defined for training data points. To generalize the transport to unseen testing data, we follow the strategy of \citet{cui2023bipartite} by applying a piecewise linear interpolation between the original and transported training scores. Specifically, given original training scores $\boldsymbol{\widehat{s}}_b^\text{train}$ and transported counterparts $\boldsymbol{\tilde{s}}_b^{\lambda,\text{train}}$, we define a piecewise interpolation map $\Psi$  that takes test scores $\boldsymbol{\widehat{s}}_b^\text{test}$ as input and outputs transported test scores $\boldsymbol{\tilde{s}}_b^{\lambda,\text{test}}$
\begin{equation}
\label{psi}
\Psi: \left(\boldsymbol{\widehat{s}}_b^\text{test};\; \boldsymbol{\widehat{s}}_b^\text{train},\; \boldsymbol{\tilde{s}}_b^{\lambda,\text{train}}\right) \longmapsto \boldsymbol{\tilde{s}}_b^{\lambda,\text{test}}.
\end{equation}
Specifically, for each test sample $i$ from group $b$ with score $\widehat{s}_{b,i}^\text{test}$, we locate the neighboring scores $(\widehat{s}_{b,i_1}^\text{train}, \tilde{s}_{b,i_1}^{\lambda,\text{train}})$ and $(\widehat{s}_{b,i_2}^\text{train}, \tilde{s}_{b,i_2}^{\lambda,\text{train}})$ from the training set such that $\widehat{s}_{b,i_1}^\text{train} \ge \widehat{s}_{b,i}^\text{test} \ge \widehat{s}_{b,i_2}^\text{train}.$
We then apply linear interpolation
\[
\tilde{s}_{b,i}^{\lambda,\text{test}} = \tilde{s}_{b,i_1}^{\lambda,\text{train}} + \left(\tilde{s}_{b,i_2}^{\lambda,\text{train}} - \tilde{s}_{b,i_1}^{\lambda,\text{train}}\right) \cdot \frac{\widehat{s}_{b,i}^\text{test} - \widehat{s}_{b,i_1}^\text{train}}{\widehat{s}_{b,i_2}^\text{train} - \widehat{s}_{b,i_1}^\text{train}}.
\]
If $\widehat{s}_{b,i}^\text{test}$ falls outside the range of training scores, we extrapolate using the closest boundary value. The full algorithm is provided in Algorithm \ref{alg:full_ppot}.

We choose to apply optimal transport to the top-$\lambda$ portion of the score distribution, as opposed to other options, e.g., randomly selecting a $\lambda$ fraction of samples. The intuition is: assuming group $b$ is disadvantaged relative to group $a$, the cross-group AUC disparity $\Delta\mathrm{xAUC}(a,b)$ arises typically because the scoring function fails to adequately distinguish positive instances from group $b$ and negative instances from group $a$. A natural approach to mitigating this issue is to increase the scores of likely-positive instances in group $b$, thereby improving their ranking relative to the negative samples in group $a$, and reducing the xAUC gap. However, since ground-truth labels are not accessible during model deployment,  we cannot directly identify and adjust only the positive instances. Instead, we approximate this objective by targeting the instances in group $b$ with the highest predicted scores, under the assumption that these represent the most confident positive predictions from the model. This top-$\lambda$ region of the score distribution serves as a proxy for true positives, allowing us to intervene in a targeted and utility-preserving manner. In contrast, randomly selecting a $\lambda$ fraction of instances would more likely include both positives and negatives. Modifying the scores of negative instances introduces unnecessary noise because such changes do not help reduce xAUC disparity but instead distort the score distribution without effectively addressing the root cause of xAUC disparity.

\paragraph{Partial AUC Case.}
Let $\boldsymbol{\hat{s}}_a^\alpha$ and $\boldsymbol{\hat{s}}_b^\alpha$ denote the subsets of top-$\alpha$ scores from $f$ for groups $a$ and $b$, respectively. To reduce the partial xAUC gap $\Delta \widehat{p\mathrm{xAUC}}(a,b;\alpha)$, we apply proportional optimal transport to  align the distribution of $\boldsymbol{\hat{s}}_b^\alpha$ to that of $\boldsymbol{\hat{s}}_a^\alpha$ parametrized by $\lambda$. Owing to the selected partial AUC objective, we anticipate that the resulting trade-off curve between partial AUC and fairness can outperform the normal classifiers.

Specifically, we define empirical distributions $\mu_a^\alpha$ and $\mu_b^\alpha$ over $\boldsymbol{\hat{s}}_a^\alpha$ and $\boldsymbol{\hat{s}}_b^\alpha$ using uniform weights. We solve the discrete optimal transport problem between $\mu_b^\alpha$ and $\mu_a^\alpha$ with a cost matrix based on squared Euclidean distance, obtaining the optimal coupling $\gamma^*$ using \eqref{eqn:optimal_transport_gamma}. Then, we apply a partial barycentric transport mapping as in \eqref{phi}
\begin{equation}
\label{eq: partial_phi}
\Phi_\alpha: \left(\boldsymbol{\hat{s}}_b^\alpha; \gamma^*, \lambda\right) \longmapsto \boldsymbol{\tilde{s}}_b^{\lambda, \alpha},
\end{equation}
where $\lambda \in [0,1]$ controls the extent of transport. By varying $\lambda$ from $0$ to $1$, we obtain a family of partially transported score vectors $\{\boldsymbol{\tilde{s}}_b^{\lambda, \alpha}\}_{\lambda\in[0,1]}$. For each $\lambda$, we compute the corresponding $(\widehat{p\mathrm{AUC}}, \Delta \widehat{p\mathrm{xAUC}})$, which collectively trace out the fairness–accuracy curve. To characterize the optimal trade-offs, we also adopt the principle of Pareto optimality by removing any point that is dominated.

The post-processed scores for unseen testing data are handled similarly as described previously. The transport map $\Phi_\alpha$ is learned based on the top-$\alpha$ scores from the training set, namely, $(\boldsymbol{\hat{s}}_a^{\alpha, \text{train}}, \boldsymbol{\hat{s}}_b^{\alpha, \text{train}})$. We then apply the learned transport to the test set using a separate linear interpolation procedure as in \eqref{psi}
\begin{equation}
\label{eq: partial_psi}
\Psi_\alpha: \left(\boldsymbol{\hat{s}}_b^{\alpha, \text{test}}; \boldsymbol{\hat{s}}_b^{\alpha, \text{train}}, \boldsymbol{\tilde{s}}_b^{\lambda, \alpha, \text{train}}\right) \longmapsto \boldsymbol{\tilde{s}}_b^{\lambda, \alpha, \text{test}}.
\end{equation}
For each $\lambda \in [0,1]$, the evaluation of partial AUC and fairness gap is conducted based on the transported test scores $\boldsymbol{\tilde{s}}_b^{\lambda, \alpha, \text{test}}$, along with the unaltered scores $\boldsymbol{\hat{s}}_a^{\alpha, \text{test}}$ from group $a$. The full algorithm is provided in Appendix Algorithm \ref{alg:partial_ppot}. 

\section{Numerical Experiments}
We empirically evaluate the performance of our proposed framework in balancing between 1) AUC-fairness and 2) partial AUC–fairness. Specifically, 
%we consider:
% \begin{itemize}
%     \item \textbf{PPOT} (Section 4): applying proportional optimal transport to the entire score distribution produced by an arbitrary baseline model, aiming to improve the AUC–fairness Pareto frontier.
%     \item \textbf{Partial-PPOT} (Section 5): first optimizing a model toward better partial AUC performance over the top-$\alpha$ region, and then applying proportional optimal transport restricted to the top-$\alpha$ scores to further improve partial AUC–fairness.
% \end{itemize}
in both cases, we use FairPOT to adjust scores. The main difference lies in whether transport is applied to all the scores, as in AUC, or the top-$\alpha$ scores, as in partial AUC. 
%
% We aim to answer the following research questions (RQs): 
% \begin{itemize}
% \item \textbf{RQ1:} Can PPOT consistently achieve a more favorable set of trade-off solutions on the AUC–fairness Pareto frontier when applied as a post-processing step in different scenarios? 
% \item \textbf{RQ2:} Compared to other fairness methods, does PPOT exhibit engineering advantages such as computational efficiency?
% \item \textbf{RQ3:} For scenarios where only the top-$\alpha$ scoring region is of practical interest, can our framework partial-PPOT achieve simultaneous improvements in both partial AUC and fairness, i.e., going beyond the AUC–fairness trade-off?
% \item  \textbf{RQ4:} How can our framework contribute to more equitable decision-making in real-world high-stakes applications such as clinical screening?
% \end{itemize}

\subsection{Datasets}
We conduct empirical experiments across three datasets: (i) synthetic data with controllable group-level bias, (ii) public benchmark datasets commonly used in fairness studies, and (iii) a real-world clinical dataset for autism diagnosis.

\paragraph{Synthetic Data.} We generate a synthetic dataset with $N = 3000$ samples and $D = 5$ continuous features, evenly split between two demographic groups $a$ (advantaged) and $b$ (disadvantaged). Features are sampled from group-specific Gaussian distributions with different means and variances: $\mathcal{N}(0.8, 1.0^2)$ for group $a$ and $\mathcal{N}(0.1, 1.0^2)$ for group $b$. Each group is assigned its own logistic scoring function with independently drawn coefficients, and intercepts are iteratively calibrated to yield positive rates of approximately 0.3 for group $a$ and 0.1 for group $b$. Binary labels are then sampled from the resulting sigmoid-transformed scores. 

\paragraph{Public Benchmark Datasets.} 
\texttt{Bank} \cite{moro2014bank} contains 16 personal and financial attributes of 45,211 individuals contacted during a marketing campaign, with the task of predicting whether the client will subscribe to a term deposit. We select \emph{age} as the sensitive attribute and define individuals aged $\leq$ 25 as the \emph{advantaged} group and those aged $>$ 25 as the \emph{disadvantaged} group. This choice is motivated by both task relevance and statistical disparity: younger individuals tend to have higher subscription rates (25.6\%) than older individuals (11.4\%), suggesting a relative advantage in this setting. Although older individuals may possess more financial stability, their lower response rate in this context justifies our designation for fairness evaluation. \texttt{COMPAS} contains 7,214 criminal recidivism records of 12 features predicting whether an individual will reoffend within two years \cite{bao2021s}. We choose sensitive attribute as \emph{gender}, with \emph{female} as disadvantaged group and \emph{male} as advantaged group. Males make up about 80.7\% of the dataset and have a higher recidivism rate (47.3\%) compared to females (35.7\%). Although females are less likely to reoffend, they are often underrepresented and may be treated unfairly in decision-making processes. 

\paragraph{Autism Clinical Data.} The clinical dataset contains 43,945 longitudinal clinical records collected between January 2015 and December 2023. Each patient record includes 21 demographic features (e.g., age, sex) and 229 diagnosis features derived from the Clinical Classifications Software (CCS) taxonomy, a standardized grouping of ICD diagnostic codes. The prediction target is a binary indicator for autism spectrum disorder (ASD) diagnosis. Due to the higher prevalence of ASD in males compared to females, there is a significant imbalance in the label distribution across gender, with the positive rate for males at 3.1\% and for females at 1.0\%. This naturally occurring divergence results in bias towards female. Therefore, we choose sensitive attribute as \emph{gender}, with \emph{female} as disadvantaged group and \emph{male} as advantaged group.
\begin{figure*}[t!]
\centering
\includegraphics[width=0.86\textwidth]{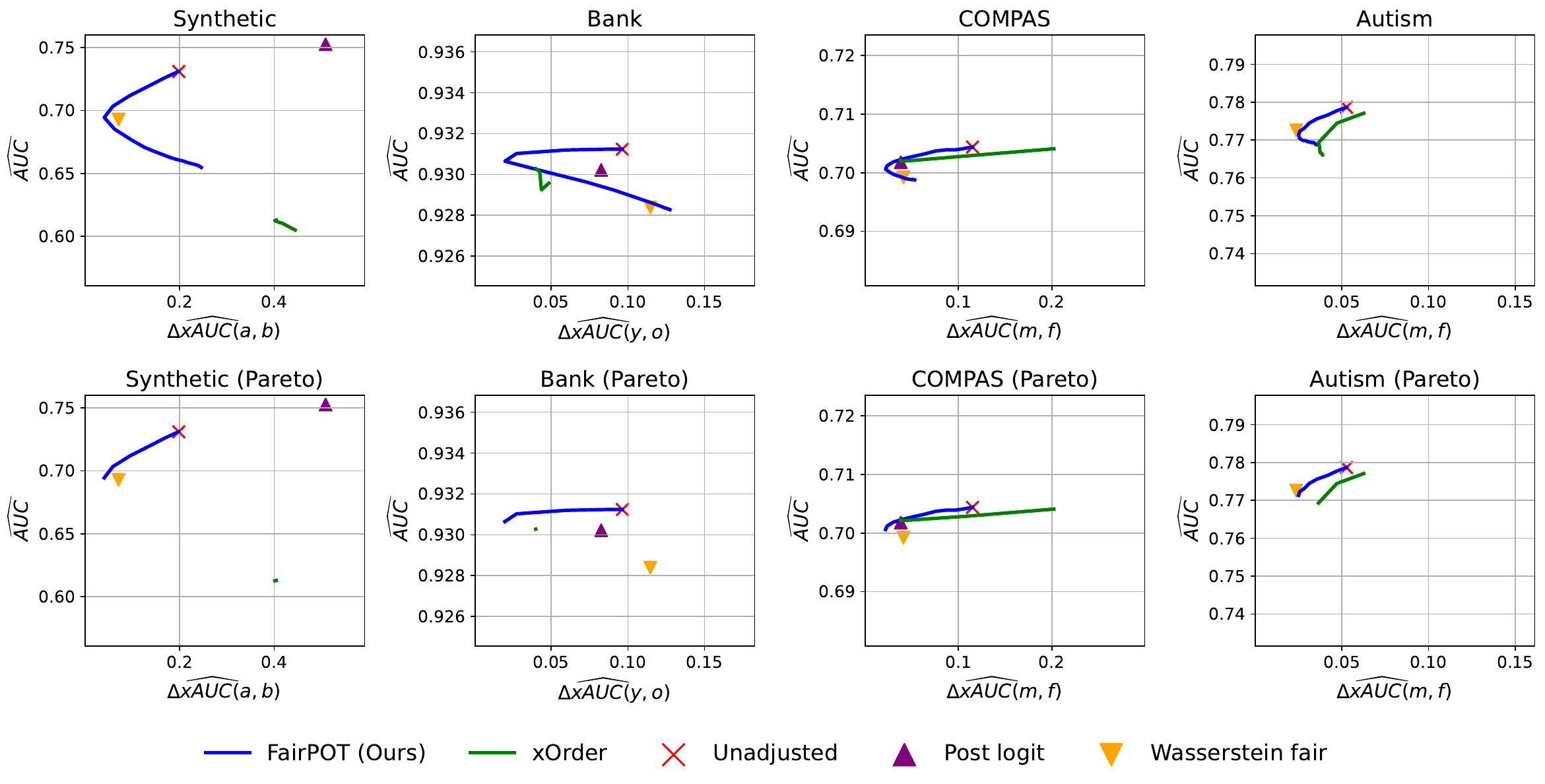} % Reduce the figure size so that it is slightly narrower than the column.
\caption{$\widehat{\mathrm{AUC}}-\Delta \widehat{\mathrm{xAUC}}$ trade-off.}
\label{fig:ppot}
\end{figure*}

\begin{figure*}[t!]
\centering
\includegraphics[width=0.86\textwidth]{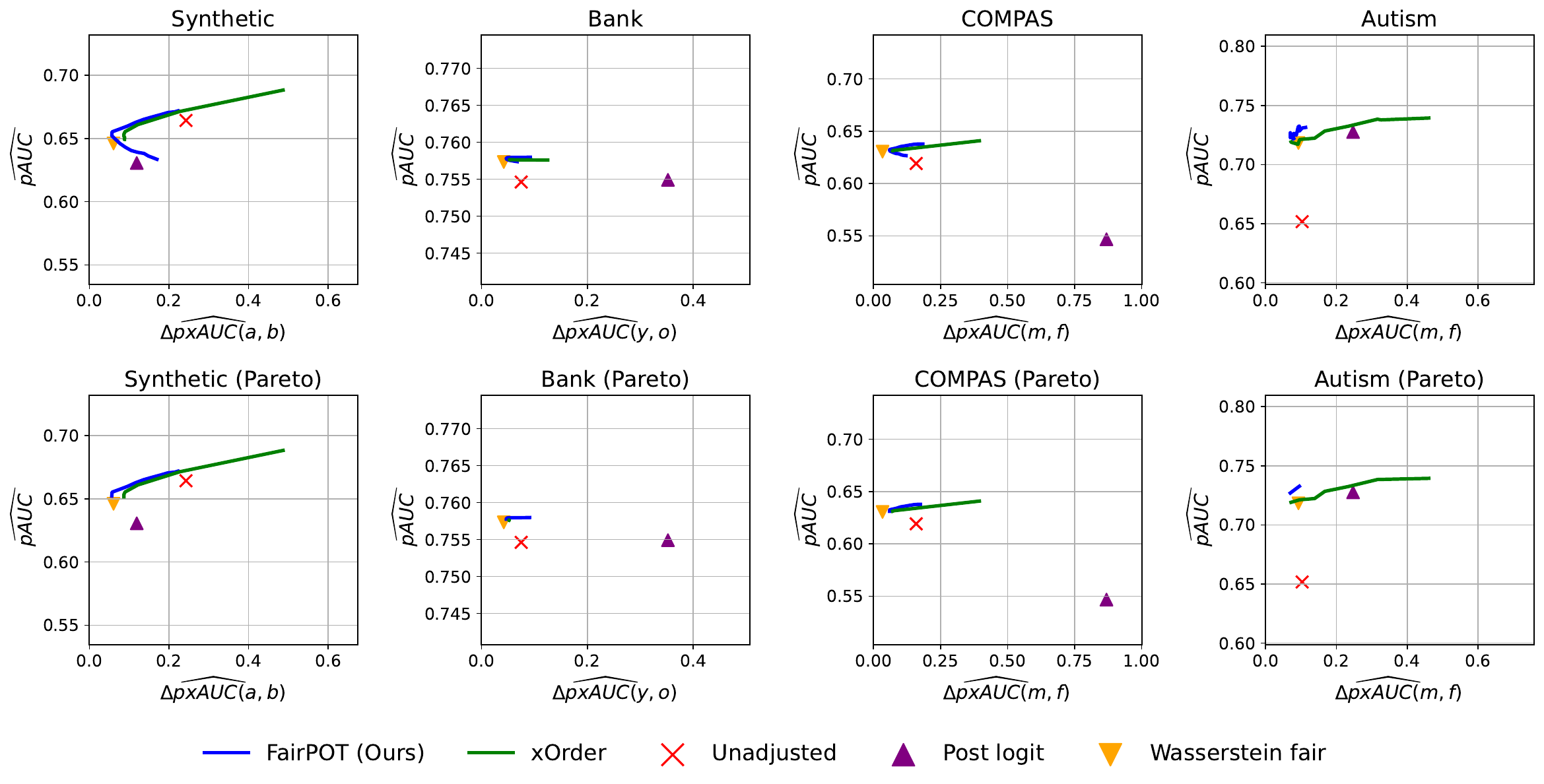} % Reduce the figure size so that it is slightly narrower than the column.
\caption{$\widehat{p\mathrm{AUC}}-\Delta \widehat{p\mathrm{xAUC}}$ trade-off. We set \(\alpha = 0.3\) for Synthetic, Bank, and COMPAS datasets, and \(\alpha = 0.1\) for Autism.}
\label{fig:partial_ppot}
\end{figure*}

\begin{table*}[htbp]
\centering
\caption{Runtime (in seconds). Each entry shows mean ± std over 20 runs.}
\label{tab:runtime}
\small
\begin{tabular}{lcccc}
\toprule
\textbf{Dataset} & \textbf{Post logit} & \textbf{Wasserstein fair}  & \textbf{xOrder} & \textbf{FairPOT (Ours)} \\
\midrule
Synthetic & 0.195 ± 0.002 & 0.011 ± 0.000 & 3.687 ± 0.141 & 0.431 ± 0.026 \\
Bank       & 0.445 ± 0.006 & 0.075 ± 0.005 & 276.700 ± 7.469 & 53.691 ± 1.975 \\
COMPAS     & 0.230 ± 0.003 & 0.023 ± 0.001 & 20.314 ± 0.354 & 2.042 ± 0.049 \\
Autism     & 1.242 ± 0.027 & 0.264 ± 0.015 & 2125.681 ± 18.857 & 360.578 ± 4.793 \\
\bottomrule
\end{tabular}
\end{table*}

\subsection{Baselines}
We evaluate the performance of FairPOT in the context of post-processing, where it is most comparable in terms of implementation and assumptions. Pre-processing and in-processing methods often require access to raw features, model gradients, or retraining, making it difficult to establish a fair and consistent comparison with our proposed FairPOT. We compare our method against several representative post-processing baselines. These methods are chosen because they are model-agnostic, require no retraining, and aim to improve score-based group fairness:
\begin{itemize}
\item \textbf{Unadjusted}: Standard XGBoost \cite{chen2016xgboost} without any fairness intervention. We choose XGBoost as the base model due to its strong empirical performance and widespread adoption in tabular data tasks. It serves as the scoring function for all methods to ensure consistent comparison.

\item \textbf{Post-logit} \cite{kallus2019fairness}: A group-specific score transformation method originally introduced in the xAUC paper. It applies a scaled sigmoid function to the scores of the disadvantaged group $b$: $\tilde{s}_b = \sigma(\alpha \cdot s_b + \beta)$, where $\sigma(\cdot)$ is the sigmoid function, $\beta$ is fixed, and $\alpha$ is a scaling parameter tuned to reduce training disparity (e.g., xAUC gap). The scores of group $a$ remain unchanged. We implement this method using the public code from \citet{cui2023bipartite}.

\item \textbf{Wasserstein Fair Post-processing} \cite{jiang2020wasserstein}: A distribution-matching method that aligns score distributions across groups via quantile matching using the Wasserstein-1 distance. Each score is mapped to its group-wise quantile index, and then pushed forward to a shared barycenter quantile. This method represents a different use of optimal transport compared to FairPOT, and is included to contrast its performance.

\item \textbf{xOrder} \cite{cui2023bipartite}: A ranking-based post-processing method that reorders scores across groups by directly optimizing a value function balancing AUC and fairness. It supports fine-grained pairwise adjustments and serves as a strong baseline for score-level fairness.
\end{itemize}

\subsection{Evaluation Protocols}
\paragraph{Global AUC Case:} (i) For the synthetic and public fairness datasets, we randomly split the data into training and test sets with a 80/20 ratio. For the clinical autism dataset, we follow task-specific cohort selection criteria to split the data. 
(ii) We first train a base classifier using \textbf{XGBoost} on the training set and obtain predicted risk scores for both the training and test sets. The predicted scores from the training data are then separated into advantaged and disadvantaged groups based on the sensitive attribute. 
(iii) For \textbf{xOrder} and \textbf{FairPOT}, we first apply the post-processing procedure to the training scores of the disadvantaged group only, using the advantaged group as a reference. The learned transformation is then generalized to the test scores of the disadvantaged group via interpolation based on quantile matching, as described in the method section. For \textbf{Post-Logit}, the learned adjustment function is directly applied to the test scores of the disadvantaged group, while the advantaged group scores remain unchanged. For \textbf{Wasserstein fair}, both the disadvantaged and advantaged group scores are jointly transformed at test time by computing a Wasserstein barycenter between their score distributions.  
(iv) Finally, for each method, we choose a set of discrete trade-off parameters $\Lambda = \{0.0, 0.1, 0.2, \dots, 1.0\}$ to generate a family of adjusted score outputs. Then we compute a trade-off curve between accuracy and fairness, i.e., AUC vs. xAUC gap, enabling comprehensive comparison of different methods under different fairness-accuracy regimes. 
(v) For synthetic and public datasets, we repeat all experiments 20 times with different random seeds and report the mean to account for variability across data splits; for Autism dataset, due to task-specific cohort selection criteria, we use a fixed test set and resample 20 times with replacement from test set for bootstrap. We also report the results with standard error in the Appendix as in Figure \ref{fig:ppot_std}.

\paragraph{Partial AUC Case: } (i) We follow a similar data splitting strategy as in the first part, randomly dividing the dataset into training and test sets (80/20 split for synthetic and public datasets, task-driven split for clinical data). 
(ii) We train a base \textbf{XGBoost} on the training set to generate preliminary risk scores. These scores are then concatenated with the original features to form an augmented feature set. We use this augmented data to train a multilayer perceptron (MLP) model using a partial AUC-oriented objective as defined in \citet{zhu2022auc, yuan2023libauc}, e.g., a differentiable surrogate loss that encourages positive instances to rank above negatives within the top-$\alpha$ portion of the global ranking \cite{zhu2022auc, yang2022algorithmic, yuan2023libauc}. This yields a baseline model whose predictions already improve partial AUC.
(iii) We then define the top-$\alpha$ region of interest based on the descending order of the optimized scores. Post-processing is performed only within this region. Specifically, we apply the proposed method and baselines to align the score distribution of the disadvantaged group in the top-\(\alpha\) training region to that of the advantaged group, which is in line with the procedures described in the global AUC case. 
(iv) For synthetic and public datasets, we repeat all experiments 20 times with different random seeds and report the mean to account for variability across data splits; for Autism dataset, due to task-specific cohort selection criteria, we use a fixed test set and resample 20 times with replacement from test set for bootstrap. We also report the results with standard error in the Appendix as in Figure \ref{fig:partial_ppot_std}.

\subsection{Results}
\paragraph{Improved Pareto Frontier for $\widehat{\mathrm{AUC}}-\Delta\widehat{\mathrm{xAUC}}$ Trade-off.} As shown in Figure~\ref{fig:ppot}, FairPOT consistently achieves a superior or comparable Pareto frontier compared to baselines including xOrder, Post-logit adjustment, and Wasserstein fair across all datasets. 
In the Synthetic, the Bank, and the COMPAS datasets, the FairPOT curve strictly dominates other methods. For every level of fairness gap, FairPOT attains a higher AUC. In the Autism dataset, FairPOT dominates the baselines in most regions. Although Wasserstein fair achieves a lower fairness gap on Autism, it causes a substantial drop in AUC and lacks the flexibility to explore trade-offs between fairness and accuracy. Notably, the unadjusted point always lies strictly inside the FairPOT frontier, confirming that FairPOT can recover the original model by selecting $\lambda=0$. Among all baselines, xOrder is the only one demonstrating some degree of fairness–accuracy trade-off. 
However, FairPOT consistently yields a better Pareto frontier than xOrder across all datasets. Moreover, we observe that the Pareto curve produced by xOrder does not always pass through the original unadjusted model performance point (i.e., the $(\widehat{\mathrm{AUC}}, \Delta\widehat{\mathrm{xAUC}})$ value before any post-processing). This suggests that xOrder is unable to fully recover the original scoring function by adjusting its tuning parameter, and may introduce greater deviations from the original model predictions compared to FairPOT.
Also, xOrder is more sensitive to its tuning parameter, making it difficult to select an appropriate configuration in practice.

\paragraph{Simultaneous Improvements in $\widehat{p\mathrm{AUC}}$ and $\Delta\widehat{p\mathrm{xAUC}}$.} As shown in Figure~\ref{fig:partial_ppot}, FairPOT in the partial AUC setting demonstrates clear regions where both partial AUC and fairness are simultaneously improved compared to the unadjusted XGBoost. In the Bank, the COMPAS, and the Autism datasets, FairPOT consistently achieves joint improvements across almost the entire curve, highlighting the potential to move beyond the traditional trade-off between partial AUC and fairness in top-$\alpha$ scenarios. On Synthetic, FairPOT also identifies regions with simultaneous gains, although the improvements are more localized. While xOrder sometimes achieves slightly higher partial AUC values compared to FairPOT, these gains often come at the expense of fairness, with fairness gaps even exceeding those of the original (unadjusted) model. Such regions are less desirable in practice, as our goal is to improve fairness without sacrificing it relative to the starting point. In contrast, FairPOT provides more balanced improvements, ensuring that fairness is at least maintained or enhanced while promoting better partial AUC among top-scoring individuals.

\paragraph{Computation Efficiency.} Experiments on Synthetic, Bank, and COMPAS datasets were conducted on a personal device (Apple M4 Pro, 24 GB RAM). Experiments on the Autism dataset were conducted on a secure remote server (Intel Xeon E5-2699 v4, 16 GB RAM), due to data access restrictions preventing local processing. As summarized in Table~\ref{tab:runtime}, FairPOT offers a practical balance between fairness improvement and computational efficiency. Compared to xOrder, FairPOT is significantly faster while achieving comparable or better fairness–accuracy trade-offs. While Post-logit adjustment and Wasserstein fair run faster, they fail to achieve good Pareto trade-offs across datasets. Thus, FairPOT strikes a favorable trade-off between runtime and effectiveness, making it a practical choice for real-world applications requiring both fairness and utility. 

\paragraph{Towards Fair and Optimal Decision Making.} Our proposed framework provides actionable insights for practical decision-making. For instance, on the Autism dataset, FairPOT in the partial AUC setting not only improves partial AUC but also reduces partial fairness disparity compared to the original XGBoost, enabling more equitable identification of high-risk individuals. In domains such as healthcare, screening, and criminal justice, where resource allocation is constrained, such improvements enable more effective and fair downstream interventions. We also conduct a more detailed analysis of how the chosen $\alpha$ can affect the decision-making for Autism data, as shown in the Appendix.

\section{Conclusion}
We propose \textit{Fair Proportional Optimal Transport} (FairPOT), a simple and flexible post-processing method to improve fairness of risk scores while preserving model performance. By adjusting only part of the scores, FairPOT allows a clear trade-off between fairness and AUC. We also extend this method to partial AUC, focusing on top-risk regions important in real-world scenarios. Experiments on multiple datasets show that FairPOT performs better than existing post-processing methods, offering a good balance between fairness and utility with low computational cost.

\section{Acknowledgements}
This work is supported in part by The Duke Endowment (TDE) under grant \#7262-SP and by the Eunice Kennedy Shriver National Institute of Child Health and Human Development (NICHD) under grant P50 HD093074.

% \section{Acknowledgments}
% \bibliography{aaai25}

\clearpage
% \appendix
\appendix
\section{Appendix}
\vspace{0.38cm}

\subsection{Appendix A. Group and Score Interpretation}
\label{sec:appendix}
In this work, we define the disadvantaged group not based on score magnitude or social identity alone, but rather in terms of model performance. Specifically, the group for which the model demonstrates worse discriminative ability in ranking positive instances above negative instances from the other group. Formally, this corresponds to lower xAUC, where the probability that a randomly chosen positive from the disadvantaged group ranks above a randomly chosen negative from the advantaged group is lower than the reverse.

This framing does not assume whether higher or lower scores are inherently advantageous. In some tasks, the disadvantaged group may exhibit higher average scores. For example, older individuals in financial risk prediction may be over-scored due to age-related biases, yet the model may still fail to correctly distinguish between true positives and negatives within this group. In other cases, the disadvantaged group may have lower average scores, as observed in recidivism prediction, where female individuals tend to receive lower risk scores, but the model may underperform in identifying true positives. FairPOT is designed to flexibly handle both types of scenarios. Since it focuses on improving relative ranking fairness through cross-group pairwise comparisons, it does not rely on assumptions about the absolute direction of scores.

% Moreover, if the application context reverses the interpretation of risk (e.g., lower scores being more desirable), our method can be trivially adapted by inverting the score scale prior to applying the transport.
\subsection{Appendix B. Algorithm for Partial AUC Case of FairPOT}
Algorithm~\ref{alg:partial_ppot} outlines the pseudocode for applying FairPOT in the partial AUC setting, where the fairness intervention is restricted to the top-scoring region of the output.
\begin{algorithm}[ht]
\caption{FairPOT: Partial AUC Case.}
\label{alg:partial_ppot}
\begin{algorithmic}[1]

\State \textbf{Input:} Predicted scores $\boldsymbol{\hat{s}}^{\text{train}}$, $\boldsymbol{\hat{s}}^{\text{test}}$; group labels $\boldsymbol{g}^{\text{train}}$, $\boldsymbol{g}^{\text{test}}$; top region parameter $\alpha$, a set of discrete trade-off parameters $\Lambda = \{\lambda_1, \lambda_2, \dots, \lambda_L\} \subseteq [0,1]$.
\State \textbf{Output:} Transported scores of top $\alpha$ region $\boldsymbol{\tilde{s}}^{\lambda, \alpha, \text{train}}$, $\boldsymbol{\tilde{s}}^{\lambda, \alpha, \text{test}}$ for each $\lambda \in \Lambda$; Pareto frontier.

\State Sort $\boldsymbol{\hat{s}}^{\text{train}}$ in descending order and take top $N_\alpha^{\text{train}}$ to obtain $\boldsymbol{\hat{s}}_b^{\alpha, \text{train}}$.
\State Sort $\boldsymbol{\hat{s}}_b^{\text{test}}$ in descending order and take top $N_\alpha^{\text{test}}$ to obtain $\boldsymbol{\hat{s}}_b^{\alpha, \text{test}}$.
\State Extract group-specific top scores $\boldsymbol{\hat{s}}_a^{\alpha, \text{train}}$, $\boldsymbol{\hat{s}}_b^{\alpha, \text{train}}$.
\State Define empirical distributions $\mu_a^{\alpha, \text{train}}$, $\mu_b^{\alpha, \text{train}}$ over top-$\alpha$ samples.
\State Solve optimal transport plan
\[
\gamma^* \gets \arg\min_{\gamma \in \Gamma(\mu_b^{\alpha, \text{train}}, \mu_a^{\alpha, \text{train}})} \sum_{i,j} \gamma_{i,j} (\hat{s}_{b,i}^{\alpha, \text{train}} - \hat{s}_{a,j}^{\alpha, \text{train}})^2.
\]
\For{each $\lambda \in \Lambda$}
    \State Transport train scores
    \[
    \boldsymbol{\tilde{s}}_b^{\lambda, \alpha, \text{train}} \gets \Phi_\alpha(\boldsymbol{\hat{s}}_b^{\alpha, \text{train}}; \gamma^*, \lambda).
    \]
    \State Map test scores
    \[
    \Psi_\alpha: \left(\boldsymbol{\hat{s}}_b^{\alpha, \text{test}}; \boldsymbol{\hat{s}}_b^{\alpha, \text{train}}, \boldsymbol{\tilde{s}}_b^{\lambda, \alpha, \text{train}}\right) \longmapsto \boldsymbol{\tilde{s}}_b^{\lambda, \alpha, \text{test}}.
    \]
    \State Construct full transported scores
    \[
    \boldsymbol{\tilde{s}}^{\lambda, \alpha, \text{train}} \gets \text{Merge}(\boldsymbol{\hat{s}}_a^{\alpha,\text{train}}, \boldsymbol{\tilde{s}}_b^{\lambda, \alpha,\text{train}}),
    \]
    \[
    \boldsymbol{\tilde{s}}^{\lambda, \alpha, \text{test}} \gets \text{Merge}(\boldsymbol{\hat{s}}_a^{\alpha,\text{test}}, \boldsymbol{\tilde{s}}_b^{\lambda, \alpha,\text{test}}).\]
    
    \State Evaluate $\widehat{p\mathrm{AUC}}$ and $\Delta\widehat{p\mathrm{xAUC}}$ on $\boldsymbol{\tilde{s}}^{\lambda, \alpha, \text{test}}$.
\EndFor

\State Identify Pareto frontier among evaluated $(\Delta\widehat{p\mathrm{xAUC}}, \widehat{p\mathrm{AUC}})$ points.

\end{algorithmic}
\end{algorithm}

\subsection{Appendix C. Implementation Details}
\paragraph{Global AUC Case.} In the global AUC setting, we use XGBoost \cite{chen2016xgboost} as the base classifier, due to its consistently strong empirical performance and widespread adoption in structured/tabular data tasks as shown in Table \ref{tab:classifiers}. While we adopt XGBoost for consistency, it is important to note that FairPOT is a model-agnostic post-processing method. This means it can be applied to predictions from any model, including black-box models, without requiring retraining or model access.

\begin{table*}[ht]
\centering
\caption{AUC for different classifiers. Each entry shows mean ± std over 20 runs.}
\begin{tabular}{lccccc}
\toprule
\textbf{Classifier} & \textbf{Synthetic} & \textbf{Bank} & \textbf{COMPAS} & \textbf{Autism} & \textbf{Average} \\
\midrule
Logistic Regression & 0.7254 $\pm$ 0.0202 & 0.9059 $\pm$ 0.0036 & 0.7207 $\pm$ 0.0134 & 0.7161 $\pm$ 0.0179 & 0.7670 \\
Random Forest       & 0.7583 $\pm$ 0.0132 & 0.9281 $\pm$ 0.0025 & 0.6587 $\pm$ 0.0117 & 0.6346 $\pm$ 0.0189 & 0.7449 \\
Gradient Boosting   & \textbf{0.7670} $\pm$ 0.0143 & 0.9248 $\pm$ 0.0027 & \textbf{0.7259} $\pm$ 0.0118 & 0.6714 $\pm$ 0.0155 & 0.7723 \\
SVM                 & 0.7363 $\pm$ 0.0136 & 0.9091 $\pm$ 0.0038 & 0.7223 $\pm$ 0.0134 & 0.6648 $\pm$ 0.0164 & 0.7581 \\
Naive Bayes         & 0.7550 $\pm$ 0.0173 & 0.8083 $\pm$ 0.0062 & 0.6950 $\pm$ 0.0131 & 0.4805 $\pm$ 0.0084 & 0.6847 \\
Decision Tree       & 0.5960 $\pm$ 0.0191 & 0.7019 $\pm$ 0.0079 & 0.6084 $\pm$ 0.0130 & 0.5253 $\pm$ 0.0102 & 0.6079 \\
KNN                 & 0.7129 $\pm$ 0.0108 & 0.8116 $\pm$ 0.0063 & 0.6608 $\pm$ 0.0125 & 0.5393 $\pm$ 0.0152 & 0.6812 \\
Extra Trees         & 0.7577 $\pm$ 0.0131 & 0.9142 $\pm$ 0.0027 & 0.6423 $\pm$ 0.0112 & 0.7113 $\pm$ 0.0270 & 0.7564 \\
XGBoost             & 0.7311 $\pm$ 0.0139 & \textbf{0.9312} $\pm$ 0.0024 & 0.7044 $\pm$ 0.0105 & \textbf{0.7731} $\pm$ 0.0143 & \textbf{0.7850} \\
\bottomrule
\label{tab:classifiers}
\end{tabular}
\end{table*}

\paragraph{Partial AUC Case.}
In the partial AUC setting, we focus on improving model performance and fairness within the top-ranked region of the score distribution. To this end, we adopt the pairwise DRO-style objective introduced by \citet{zhu2022auc}, which efficiently approximates partial AUC. The optimization problem is defined as
$$
\min _{\mathbf{w}} \frac{1}{n_{+}} \sum_{\mathbf{x}_i \in \mathcal{S}_{+}} \lambda \log \left(\frac{1}{n_{-}} \sum_{\mathbf{x}_j \in \mathcal{S}_{-}} \exp \left(\frac{L\left(\mathbf{w} ; \mathbf{x}_i, \mathbf{x}_j\right)}{\lambda}\right)\right)
$$
Here, $\mathcal{S}_+$ and $\mathcal{S}_-$ represent the sets of positive and negative instances, and $L(\mathbf{w}; \mathbf{x}_i, \mathbf{x}_j)$ is a pairwise surrogate loss such as squared hinge loss. This objective prioritizes hard positive-negative pairs and concentrates learning in the top-risk region.

Additionally, we improve optimization by incorporating predicted scores from XGBoost as an extra feature input. Table \ref{tab: pauc_compare} compares the performance of:

(a) unadjusted XGBoost scores,

(b) partial AUC optimization using only original features, and

(c) partial AUC optimization using both original features and XGBoost scores.

\begin{table}[ht]
\centering
\caption{pAUC evaluated from different method.}
\begin{tabular}{lcccc}
\toprule
\textbf{Method} & \textbf{Synthetic} & \textbf{Bank} & \textbf{COMPAS} & \textbf{Autism} \\
\midrule
(a) & 0.6642 & 0.7546 & 0.6171 & 0.633 \\
(b) & \textbf{0.6798} & 0.7544 & 0.6153 & 0.573 \\
(c) & 0.6740 & \textbf{0.7584} & \textbf{0.6383} & \textbf{0.749} \\
\bottomrule
\end{tabular}
\label{tab: pauc_compare}
\end{table}

We apply fairness intervention on both global and partial AUC settings. However, we only optimize the model toward partial AUC, and not global AUC. This is because partial AUC optimization is more practical: in the partial setting, we can explicitly select instances from the top-$\alpha$ region to guide learning. In contrast, global AUC optimization requires reordering all instance pairs across the entire score distribution, which is algorithmically more challenging.

\subsection{Appendix D. Extended Result Analysis}
\begin{figure*}[t!]
\centering
\includegraphics[width=\textwidth]{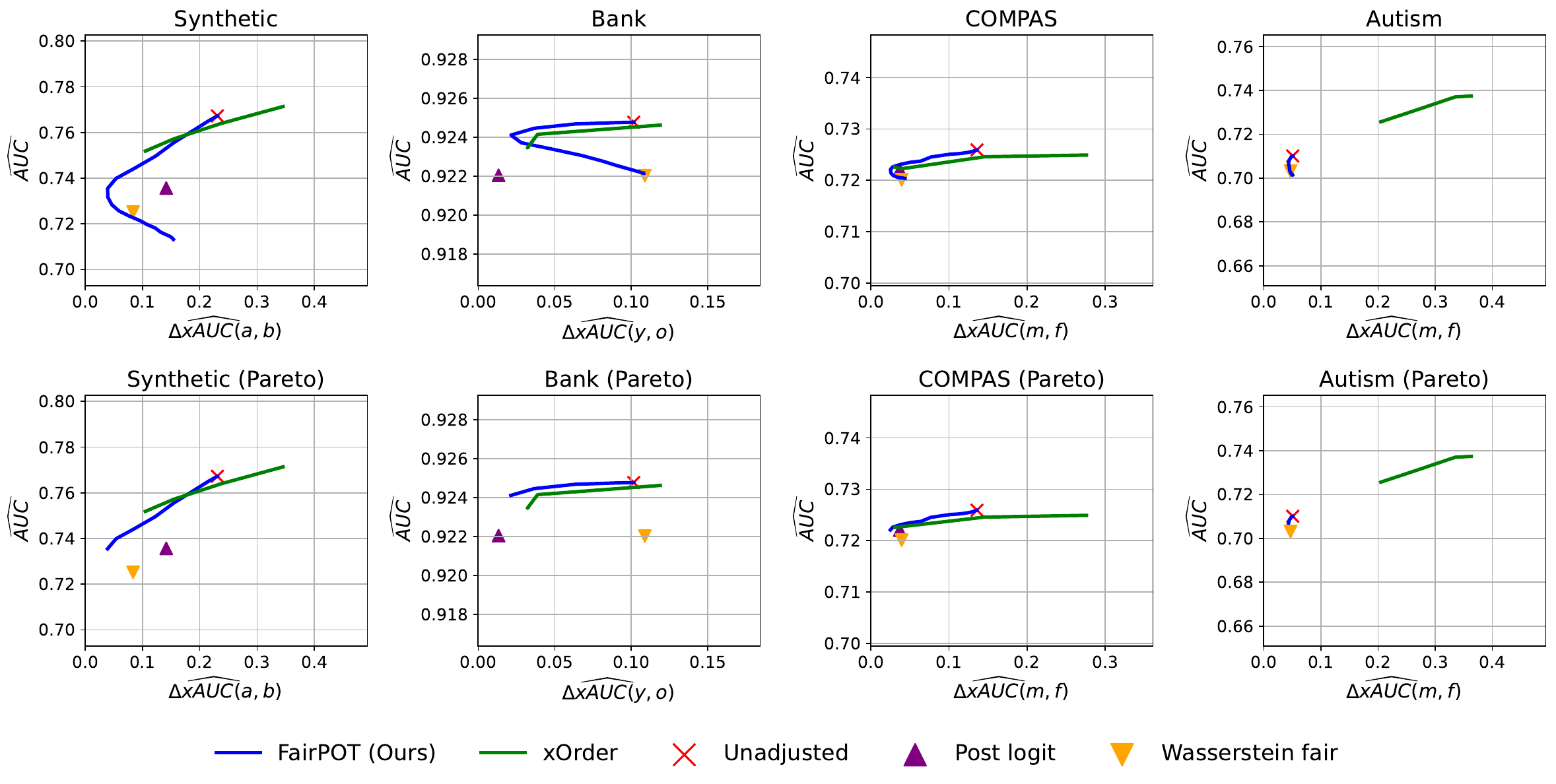} % Reduce the figure size so that it is slightly narrower than the column.
\caption{$\widehat{\mathrm{AUC}}-\Delta \widehat{\mathrm{xAUC}}$ trade-off (Gradient Boosting).}
\label{fig:ppot_gb}
\end{figure*}

\begin{figure*}[t!]
\centering
\includegraphics[width=\textwidth]{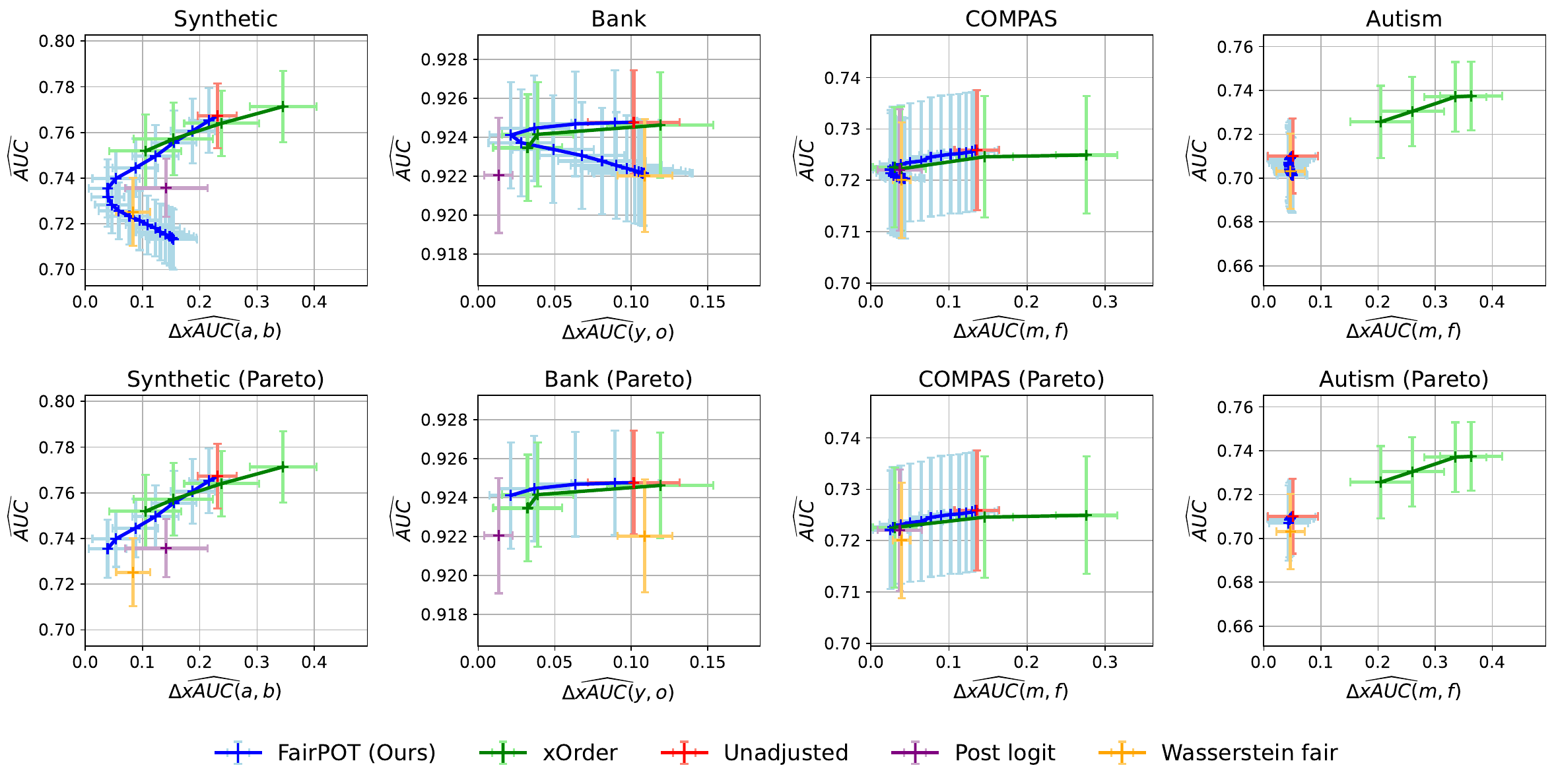} % Reduce the figure size so that it is slightly narrower than the column.
\caption{$\widehat{\mathrm{AUC}}-\Delta \widehat{\mathrm{xAUC}}$ trade-off (Gradient Boosting; mean with std bar).}
\label{fig:ppot_gb_std}
\end{figure*}

\begin{figure*}[t!]
\centering
\includegraphics[width=\textwidth]{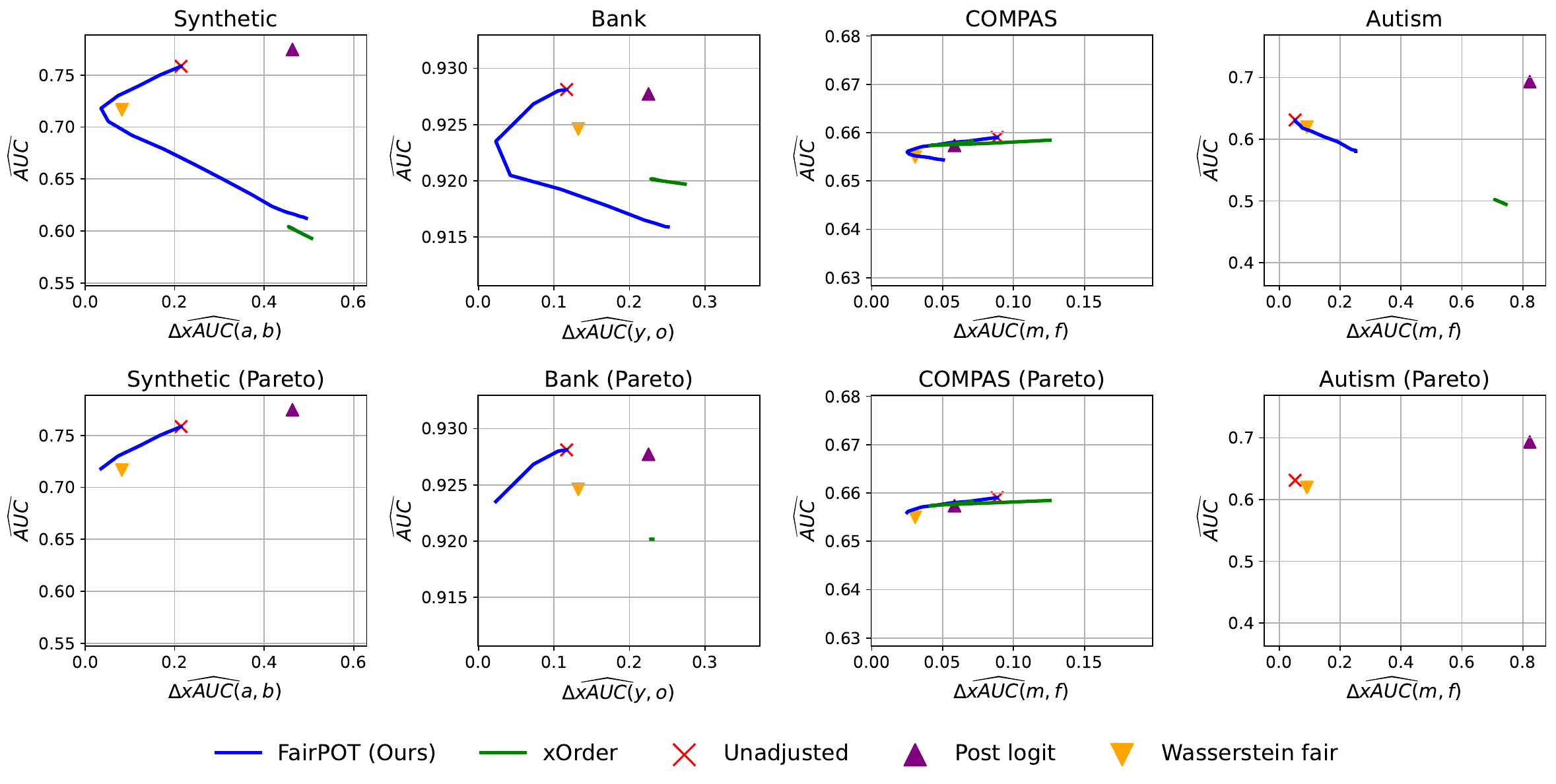} % Reduce the figure size so that it is slightly narrower than the column.
\caption{$\widehat{\mathrm{AUC}}-\Delta \widehat{\mathrm{xAUC}}$ trade-off (Random Forest).}
\label{fig:ppot_rf}
\end{figure*}

\begin{figure*}[t!]
\centering
\includegraphics[width=\textwidth]{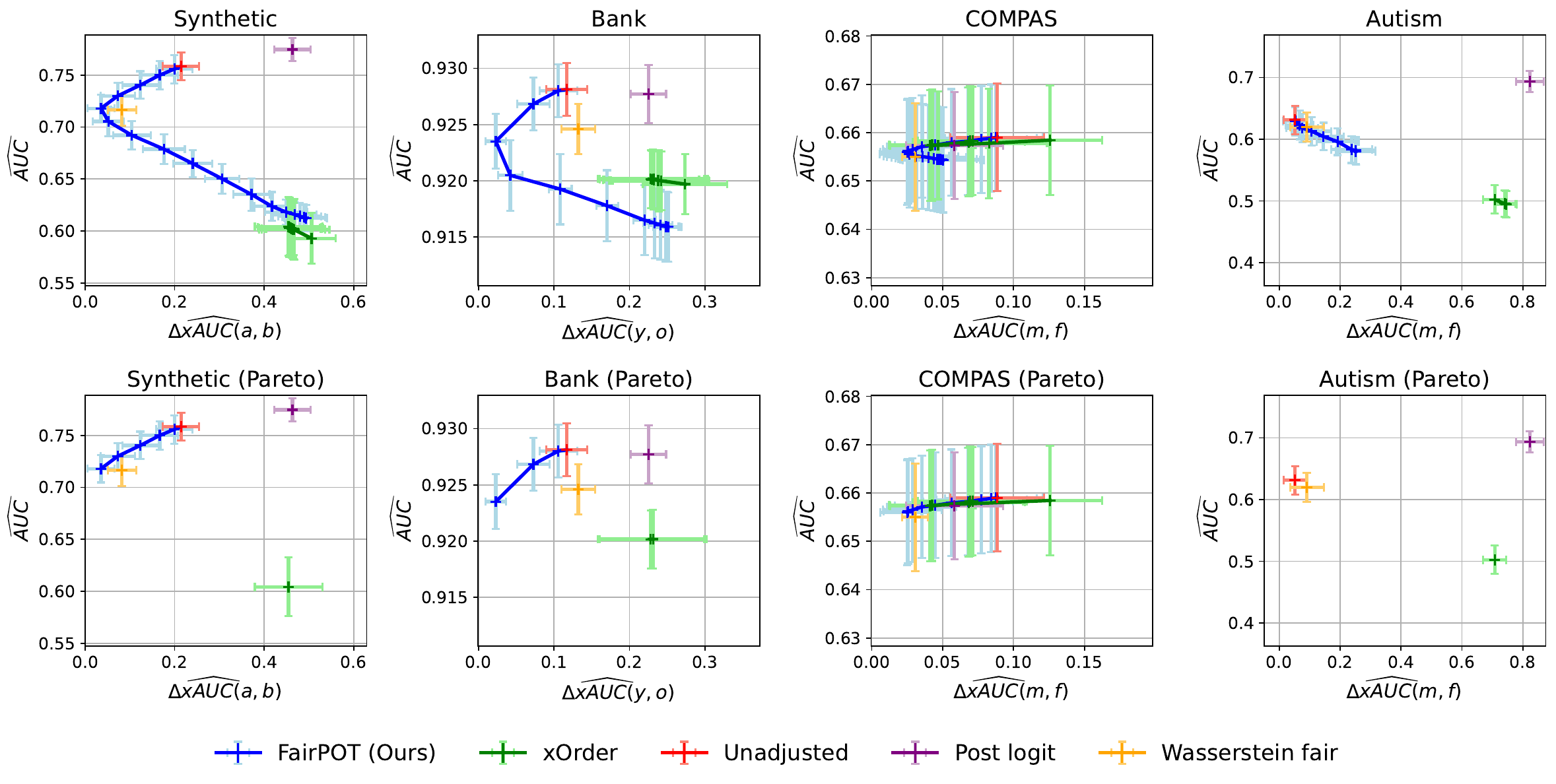} % Reduce the figure size so that it is slightly narrower than the column.
\caption{$\widehat{\mathrm{AUC}}-\Delta \widehat{\mathrm{xAUC}}$ trade-off (Random Forest; mean with std bar).}
\label{fig:ppot_rf_std}
\end{figure*}

To further demonstrate the model-agnostic nature of FairPOT, we apply it to additional classifiers—Gradient Boosting (Figures~\ref{fig:ppot_gb}, \ref{fig:ppot_gb_std}) and Random Forest (Figures~\ref{fig:ppot_rf}, \ref{fig:ppot_rf_std}). For Gradient Boosting, FairPOT achieves Pareto-dominant or comparable trade-offs across datasets. On Synthetic and COMPAS, FairPOT consistently outperforms Wasserstein Fair and Post-logit, and matches or surpasses xOrder. In the Bank dataset, Post-logit shows lower fairness disparity but with a notable drop in AUC and no ability to trade off. On Autism, xOrder fails to improve fairness, while FairPOT retains flexibility to balance fairness and accuracy. For Random Forest, FairPOT again shows strong performance. On Synthetic, it clearly outperforms all baselines, while Post-logit increases unfairness. On Bank, FairPOT dominates. On COMPAS, it achieves both improved fairness and broader control over the fairness–accuracy balance. In the Autism dataset, where the unadjusted model already yields near-parity ($\Delta\mathrm{xAUC}\approx0.01$), most baselines degrade fairness, whereas FairPOT is able to recover the original outputs with appropriately chosen $\lambda$. Overall, these results confirm the effectiveness and stability of FairPOT across different training models, supporting its practical value as a flexible and efficient post-processing solution.

\begin{figure*}[t!]
\centering
\includegraphics[width=\textwidth]{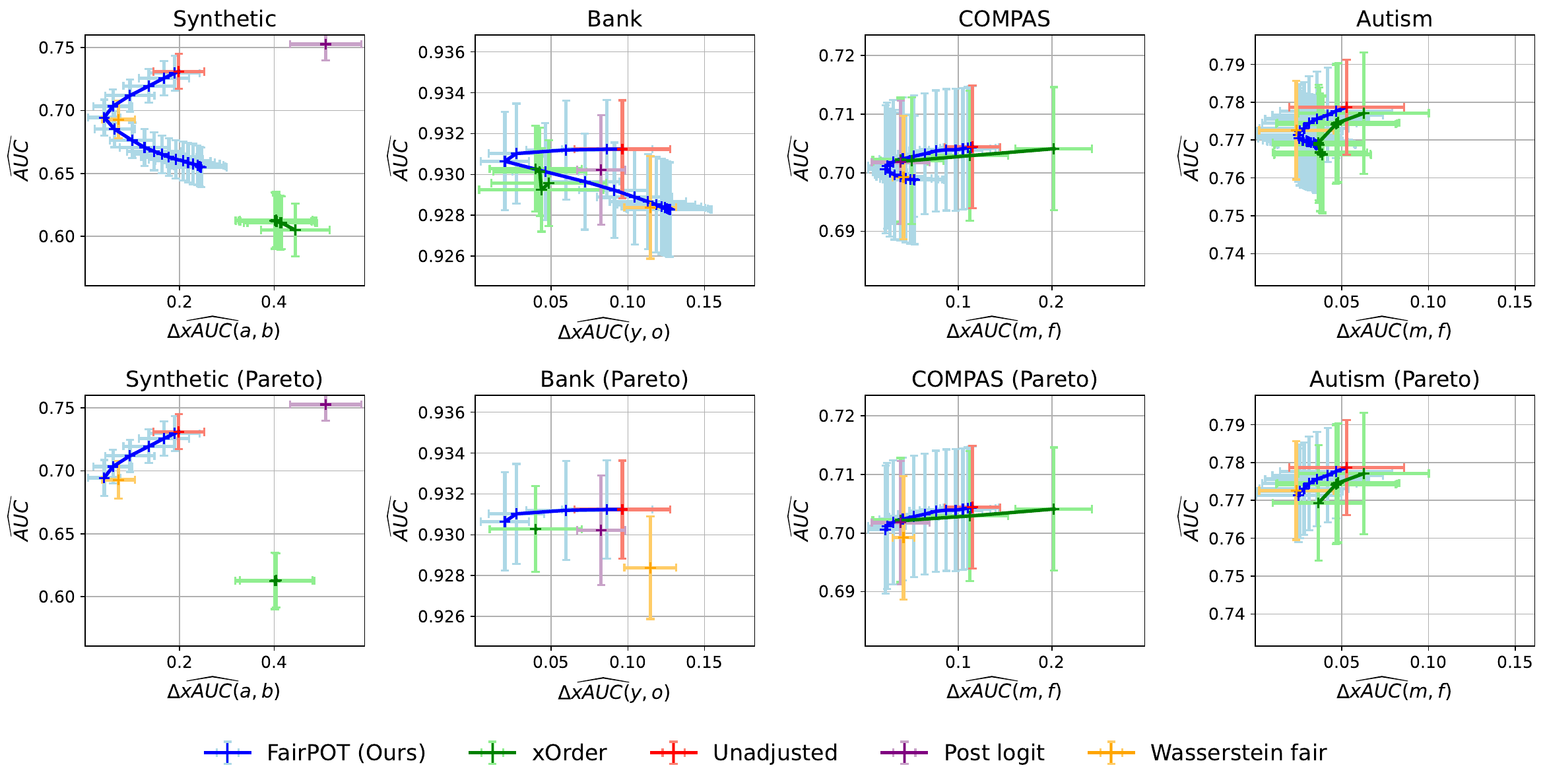} % Reduce the figure size so that it is slightly narrower than the column.
\caption{$\widehat{\mathrm{AUC}}-\Delta \widehat{\mathrm{xAUC}}$ trade-off (mean with std bar).}
\label{fig:ppot_std}
\end{figure*}

\begin{figure*}[t!]
\centering
\includegraphics[width=\textwidth]{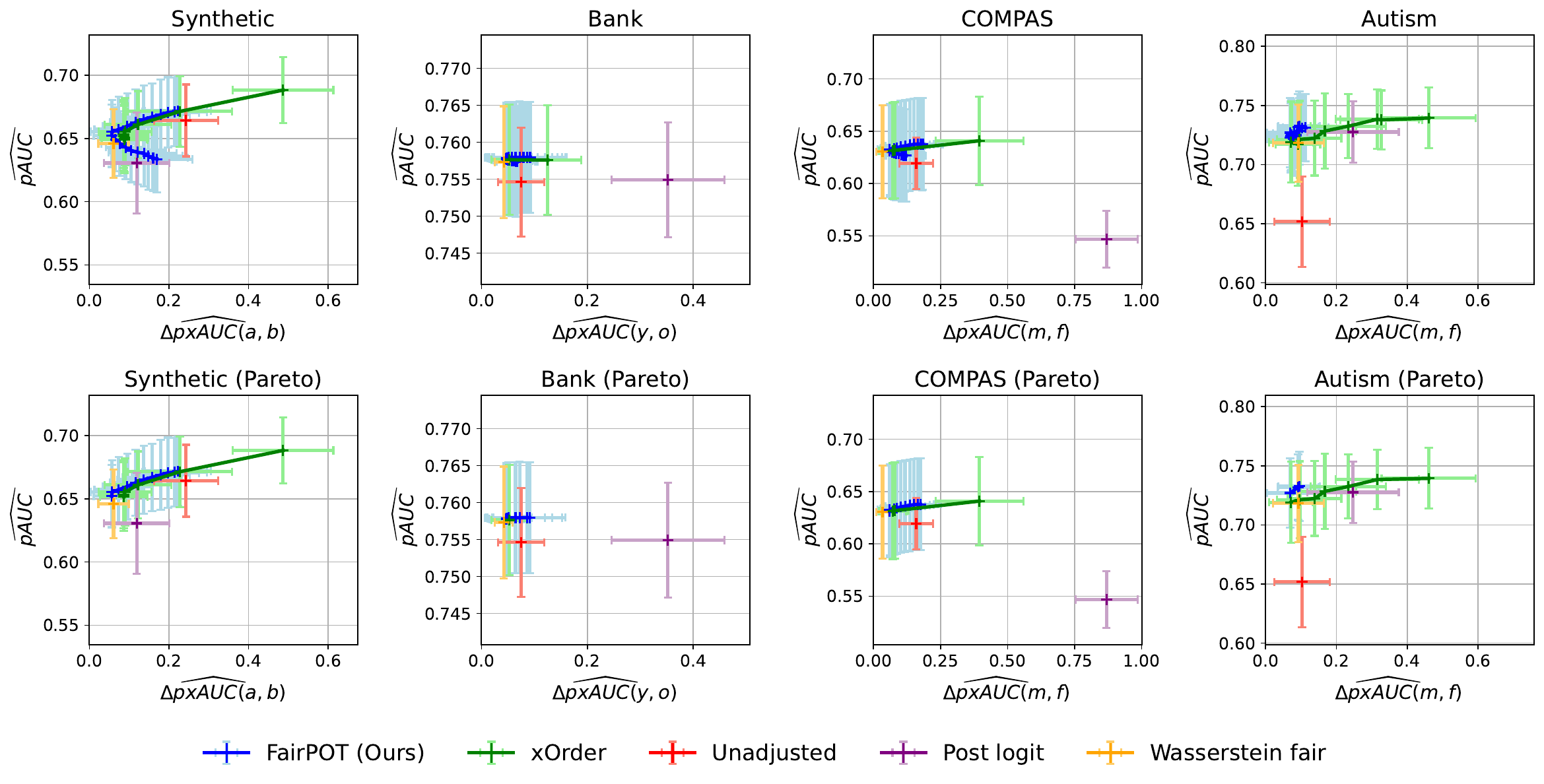} % Reduce the figure size so that it is slightly narrower than the column.
\caption{$\widehat{p\mathrm{AUC}}-\Delta \widehat{p\mathrm{xAUC}}$ trade-off (mean with std bar). We set \(\alpha = 0.3\) for Synthetic, Bank, and COMPAS datasets, and \(\alpha = 0.1\) for Autism.}
\label{fig:partial_ppot_std}
\end{figure*}

Figure~\ref{fig:ppot_std} and Figure~\ref{fig:partial_ppot_std} present the mean and standard error corresponding to the results shown in Figure~\ref{fig:ppot} and Figure~\ref{fig:partial_ppot}, respectively.

\begin{figure*}[t!]
\centering
\includegraphics[width=0.7\textwidth]{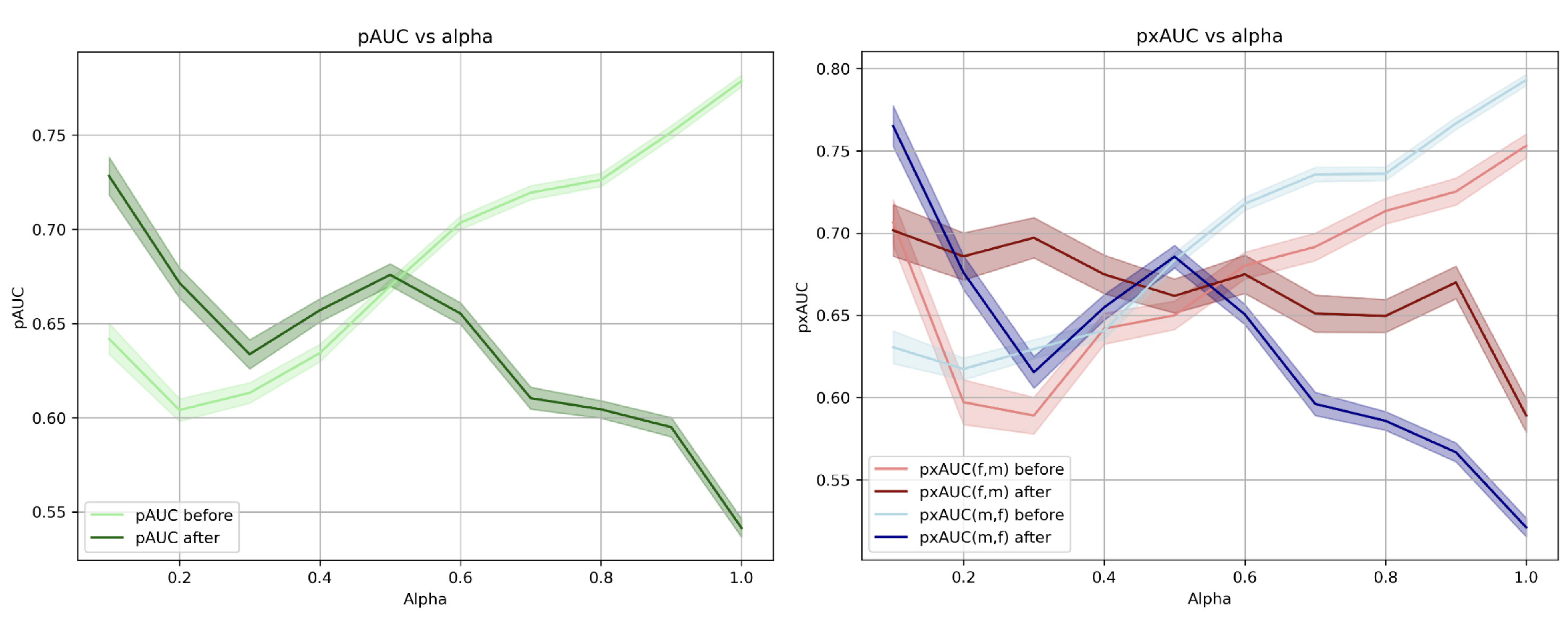} % Reduce the figure size so that it is slightly narrower than the column.
\caption{$p\mathrm{AUC}$ vs $\alpha$, $p\mathrm{xAUC}$ vs $\alpha$ for Autism Data.}
\label{fig:bootstrap_alpha}
\end{figure*}

Figure~\ref{fig:bootstrap_alpha} illustrates how $\widehat{p\mathrm{AUC}}$ and $\widehat{p\mathrm{xAUC}}$ vary with $\alpha$, both with and without applying partial AUC optimization as described in the partial AUC case of the evaluation protocols. From Figure~\ref{fig:bootstrap_alpha}, we observe that under the original XGBoost model, both $\widehat{p\mathrm{AUC}}$ and $\widehat{p\mathrm{xAUC}}$ tend to increase as $\alpha$ grows.  
This indicates that the model performs relatively worse among higher-risk individuals (i.e., those with higher predicted scores), which is undesirable in practical settings, especially for the Autism dataset where the positive rate is only around 2\%. After applying partial AUC optimization, however, this trend is reversed: both $\widehat{p\mathrm{AUC}}$ and $\widehat{p\mathrm{xAUC}}$ decrease as $\alpha$ increases. This suggests that the ability of the model to distinguish high-risk individuals has been effectively enhanced, precisely in the regions that matter most for clinical decision-making. Such improvement is particularly important in low-prevalence conditions like autism, where better score calibration in the high-risk region provides stronger support for thresholding decisions and offers greater flexibility for achieving fairness in downstream interventions.

Note that in this experiment, we apply the partial AUC optimization procedure as described in the partial AUC evaluation protocol. Therefore, the model used here differs from the base model used in the main experiments (which directly employ XGBoost without partial AUC tuning). This explains why the results for $\alpha = 1$ in Figure~\ref{fig:bootstrap_alpha} differ from the global AUC results shown earlier. Although \(\alpha = 1\) formally recovers the global AUC region, the optimization objective and resulting model are different, which leads to the observed discrepancy.

This highlights an important distinction: while the global AUC case evaluates the overall discrimination ability of base model, the partial AUC case (even with \(\alpha = 1\)) focuses on optimizing a specific risk region. Understanding this distinction helps unify the interpretation of FairPOT under different evaluation settings.

\subsection{Appendix E. Clarifying Suboptimal $\lambda$ Configurations in Trade-off Curves}
\label{app:lambda-suboptimal}

As shown in Figures~\ref{fig:ppot} and \ref{fig:partial_ppot}, we visualize the full performance curves of FairPOT across different values of $\lambda \in [0,1]$ to illustrate the trade-off between fairness and accuracy. In this process, certain $\lambda$ values may yield both lower $\widehat{\mathrm{AUC}}$ and higher $\Delta \widehat{\mathrm{xAUC}}$ (or their partial counterparts) compared to the unadjusted model ($\lambda = 0$). While these points may appear undesirable, we emphasize that they are included purely for completeness and transparency of the entire trade-off curve.

In practice, we do not select such suboptimal configurations. As discussed in Section~4, FairPOT consistently yields a superior or comparable Pareto frontier compared to baseline methods, and always includes the original model as a special case with $\lambda = 0$. To accommodate diverse application needs, we present the full Pareto-optimal region, which represents the set of non-dominated trade-offs between fairness and accuracy. The final selection of a specific $\lambda$ value should be made based on the decision maker’s preference over fairness and utility, depending on the context of deployment.

\subsection{Appendix F. Ablation: Switching Advantaged Group and Disadvantaged Group}
\label{app:ablation-shift}
Our main method applies FairPOT to shift the scores of the disadvantaged group $b$ toward the support of the advantaged group $a$, followed by interpolation to test data. A natural question arises: does the direction of transport matter? That is, what if we switch group $a$ and group $b$?

\begin{figure*}[h]
\centering
\includegraphics[width=\linewidth]{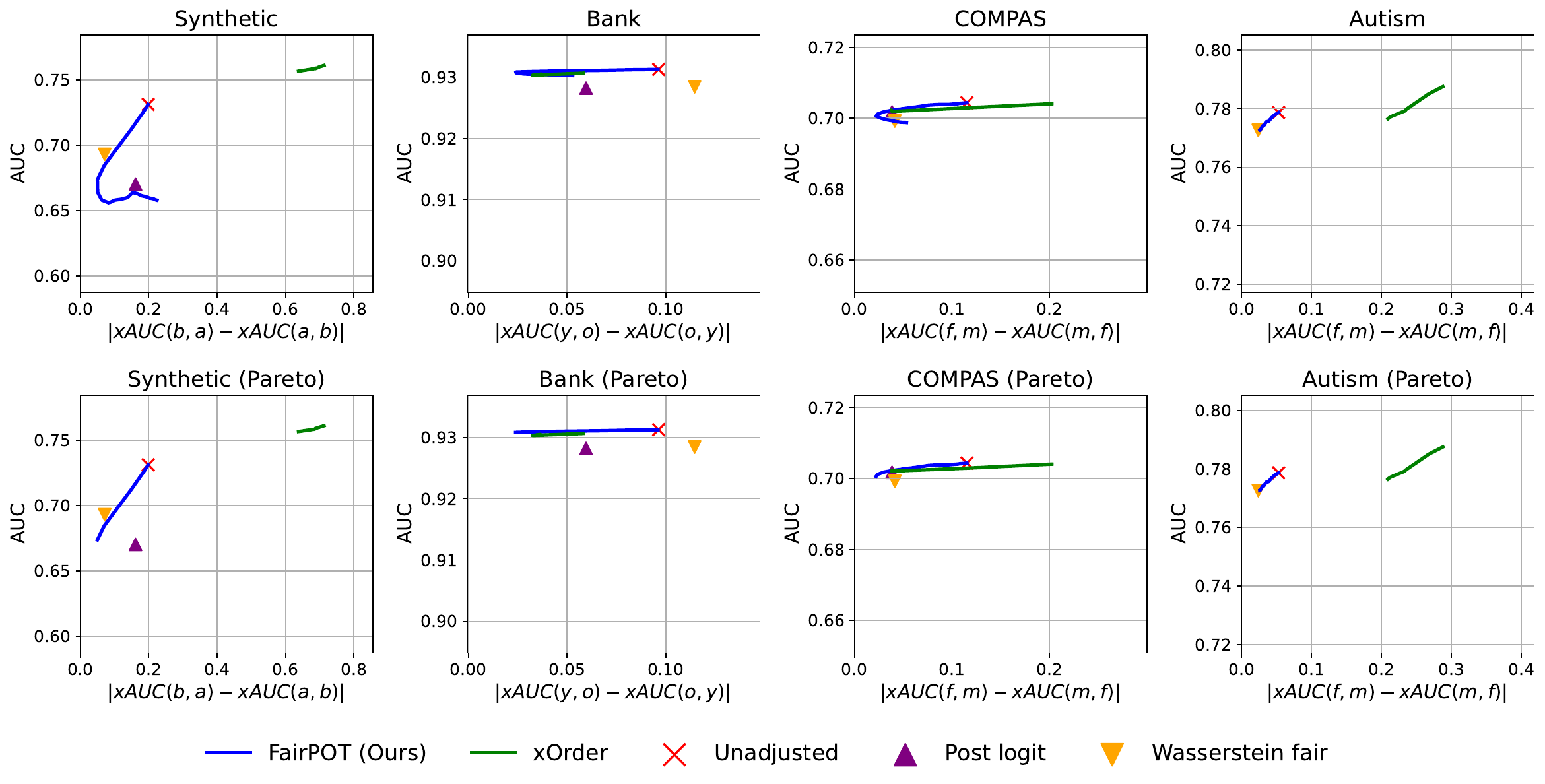}
\caption{$\widehat{\mathrm{AUC}}-\Delta \widehat{\mathrm{xAUC}}$ trade-off for ablation study.}
\label{fig:ablation-shift}
\end{figure*}

\begin{figure*}[h]
\centering
\includegraphics[width=\linewidth]{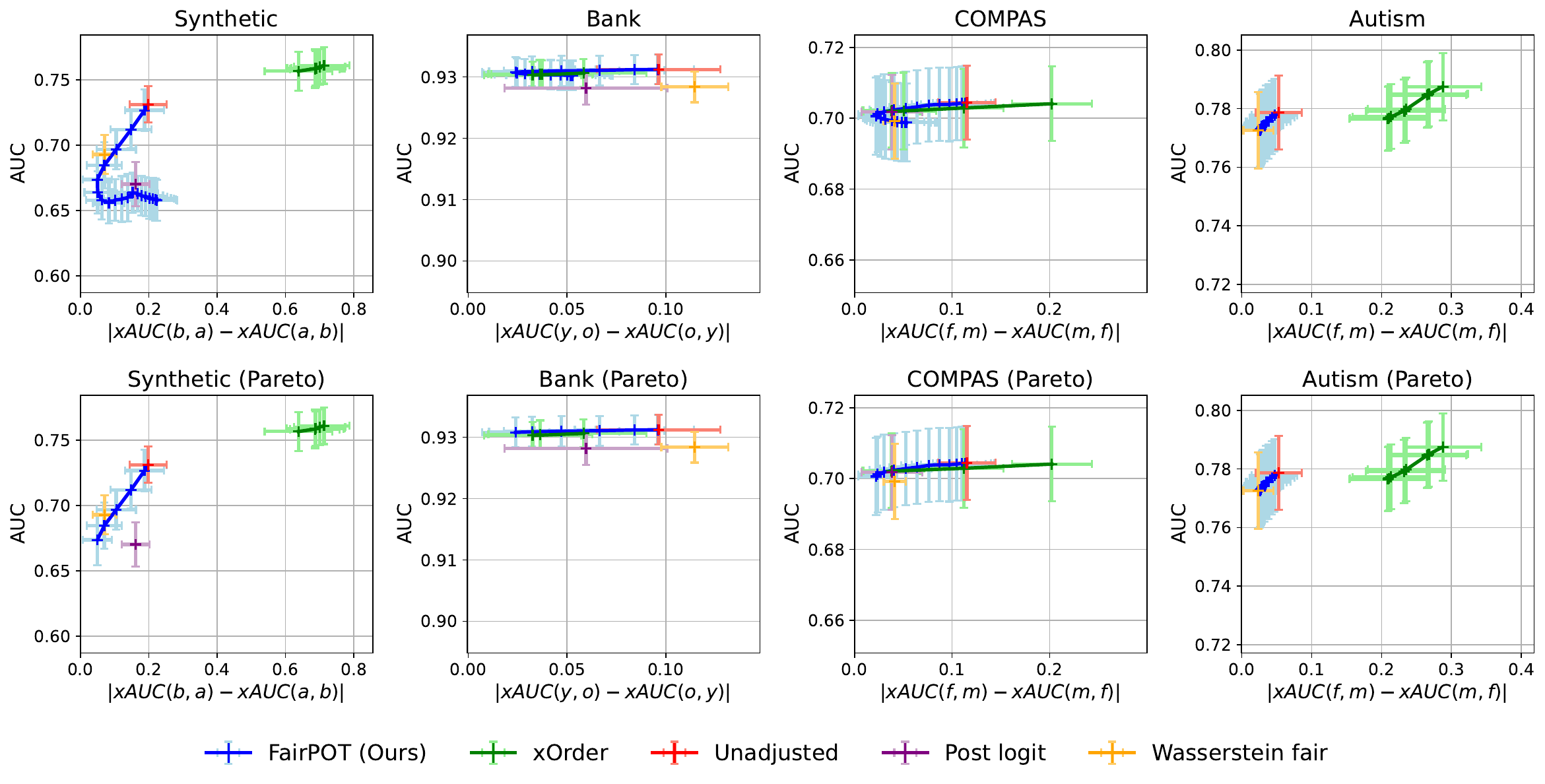}
\caption{$\widehat{\mathrm{AUC}}-\Delta \widehat{\mathrm{xAUC}}$ trade-off for ablation study  (mean with std bar).}
\label{fig:ablation-shift-std}
\end{figure*}

To answer this, we repeat our method on all four datasets (Synthetic, Bank, COMPAS, and Autism), reversing the transport direction. Specifically, we shift the training scores of group $a$ toward the support of group $b$, while keeping the test-time interpolation procedure unchanged. From Figure~\ref{fig:ablation-shift} and Figure~\ref{fig:ablation-shift-std}, we observe negligible differences between the two transport directions in terms of the $\widehat{\mathrm{AUC}}$–$\Delta \widehat{\mathrm{xAUC}}$ trade-off, which suggests that either direction yields comparable fairness–accuracy outcomes. We adopt the convention of shifting the disadvantaged group in our main method to align with fairness literature, which emphasizes mitigating harm for disadvantaged populations.

% \begin{figure*}[t!]
% \centering
% \includegraphics[width=\textwidth]{LaTeX/PPOT_lr.pdf} % Reduce the figure size so that it is slightly narrower than the column.
% \caption{$\widehat{\mathrm{AUC}}-\Delta \widehat{\mathrm{xAUC}}$ Trade-off (LogisticRegression as Unadjusted).}
% \label{fig:ppot_std}
% \end{figure*}

% \begin{figure*}[t!]
% \centering
% \includegraphics[width=\textwidth]{LaTeX/PPOT_std_lr.pdf} % Reduce the figure size so that it is slightly narrower than the column.
% \caption{$\widehat{\mathrm{AUC}}-\Delta \widehat{\mathrm{xAUC}}$ Trade-off (mean with std bar, LogisticRegression as Unadjusted).}
% \label{fig:partial_ppot_std}
% \end{figure*}

\end{document}